\documentclass[sigconf]{aamas}  

\usepackage{booktabs}
\usepackage[utf8]{inputenc} 
\usepackage[T1]{fontenc}    
\usepackage{hyperref}       
\usepackage{url}            
\usepackage{booktabs}       
\usepackage{amsfonts}       
\usepackage{nicefrac}       
\usepackage{microtype}      

\usepackage{times}
\usepackage{url}
\usepackage{graphicx}
\usepackage{subcaption}
\usepackage{booktabs}
\usepackage{float, amsmath, mathtools}
\usepackage{amsfonts}
\usepackage{color}
\usepackage{array,multirow,balance}
\usepackage{wrapfig}

\def\vct#1{\mbox{\boldmath $#1$}}
\def\eg{{\it e.g.}}

\def\ie{{\it i.e.}}

\def\etc{{\it etc}}

\usepackage{bm}
\renewcommand{\vec}[1]{\mbox{\bm{$#1$}}}

\newcommand{\mgp}{\textsc{MGpi}}

\usepackage{flushend}
\setcopyright{ifaamas}  
\acmDOI{doi}  
\acmISBN{}  
\acmConference[AAMAS'20]{Published:\;\; Proc.\@ of the 19th International Conference on Autonomous Agents and Multiagent Systems (AAMAS 2020), B.~An, N.~Yorke-Smith, A.~El~Fallah~Seghrouchni, G.~Sukthankar (eds.)}{May 2020}{Auckland, New Zealand}  
\acmYear{2020}  
\copyrightyear{2020}  
\acmPrice{}  

\begin{document}

\title{\mgp{}: A Computational Model of \\Multiagent Group Perception and Interaction}  



\author{Navyata Sanghvi}
\affiliation{
\institution{Carnegie Mellon University}
\city{Pittsburgh} 
\state{Pennsylvania} 
}
\email{navyatasanghvi@cmu.edu}

\author{Ryo Yonetani}
\authornote{Ryo Yonetani is currently at OMRON SINIC X, Tokyo, Japan.}
\affiliation{
\institution{Carnegie Mellon University}
\city{Pittsburgh} 
\state{Pennsylvania} 
}
\email{ryo.yonetani@sinicx.com}

\author{Kris Kitani}
\affiliation{
\institution{Carnegie Mellon University}
\city{Pittsburgh} 
\state{Pennsylvania} 
}
\email{kkitani@cs.cmu.edu}

\begin{abstract}
Toward enabling next-generation robots capable of socially intelligent interaction with humans, we present a {\bf computational model} of interactions in a social environment of multiple agents and multiple groups. The Multiagent Group Perception and Interaction (\mgp{}) network is a deep neural network that predicts the appropriate social action to execute in a group conversation (\emph{e.g.}, speak, listen, respond, leave), taking into account neighbors' observable features (\emph{e.g.,} location of people, gaze orientation, distraction, \emph{etc}.). A central component of \mgp{} is the Kinesic-Proxemic-Message (KPM) gate, that performs social signal gating to extract important information from a group conversation. In particular, KPM gate filters incoming social cues from nearby agents by observing their body gestures (kinesics) and spatial behavior (proxemics). The \mgp{} network and its KPM gate are learned via imitation learning, using demonstrations from our designed {\bf{social interaction simulator}}. Further, we demonstrate the efficacy of the KPM gate as a social attention mechanism, achieving state-of-the-art performance on the task of {\bf{group identification}} without using explicit group annotations, layout assumptions, or manually chosen parameters.
\end{abstract}

\keywords{Social agent models; Socially interactive agents; Agent-based analysis of human interactions; Social group identification; Multiagent learning; Learning from demonstrations.}  

\maketitle

\section{Introduction}\label{sec:intro}

In order to develop next-generation robots that can interact socially with humans, there is first a need for expressive computational models that can encode social intelligence \cite{albrecht2006social, goleman2007social, dautenhahn2007socially, vinciarelli2009social}. Broadly, an agent's social intelligence is its ability to understand and respond appropriately to others, \ie, its: (1) social perception and (2) social interaction management skill. Social perception is the ability to analyze other agents' social signals including non-verbal behavioral information such as facial or postural expressions \cite{sanghvi2011automatic,mehrabian1967inference}, physical distance ({\em{proxemics}}) \cite{edward1966hall} and body gestures ({\em{kinesics}}) \cite{birdwhistell1952introduction}. Social interaction management is the ability to take appropriate action in response to information retrieved via social perception.

\begin{figure}[t!]
\vspace{2mm}
    \centering
    \includegraphics[width=0.45\textwidth]{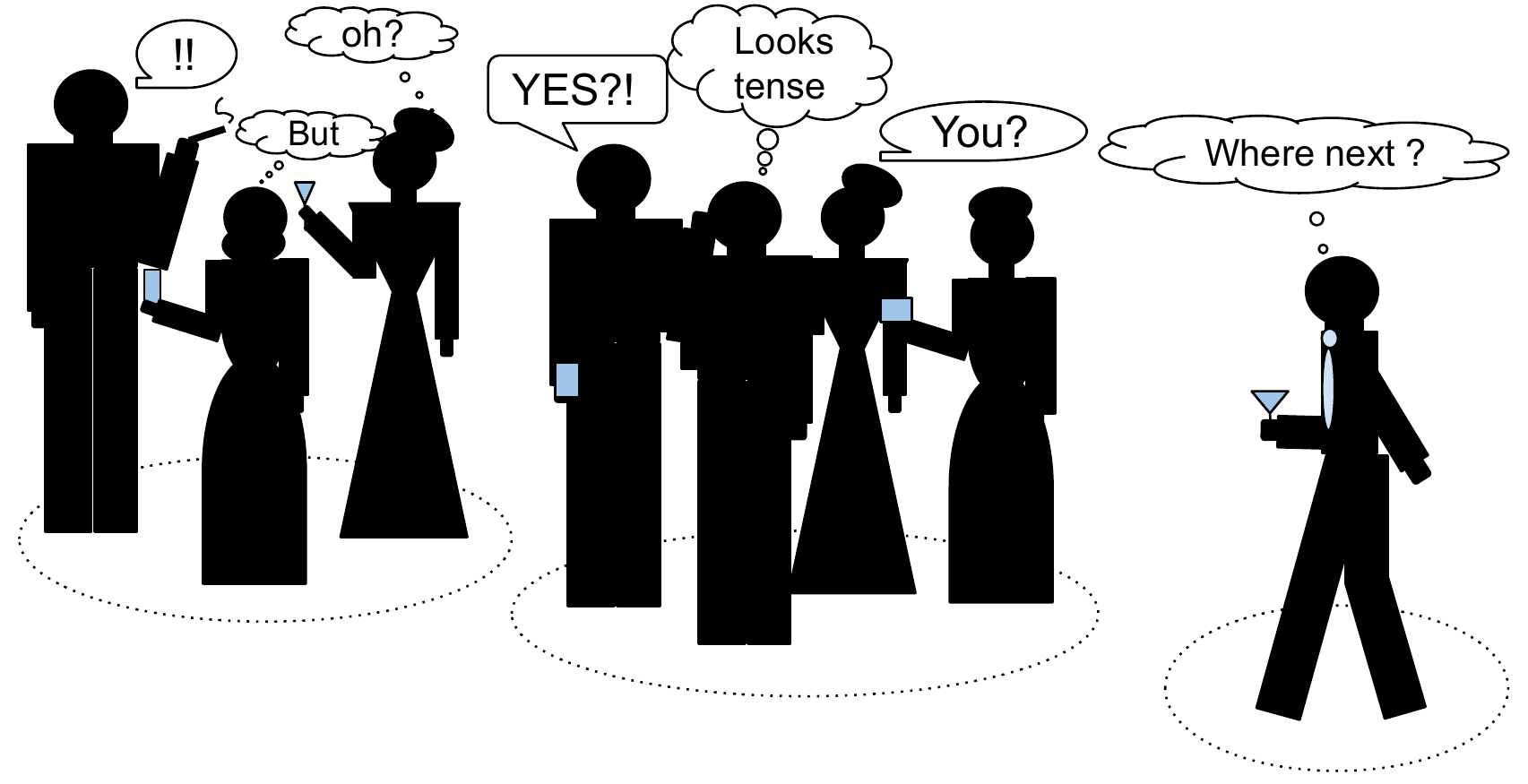}
    \includegraphics[height=27mm]{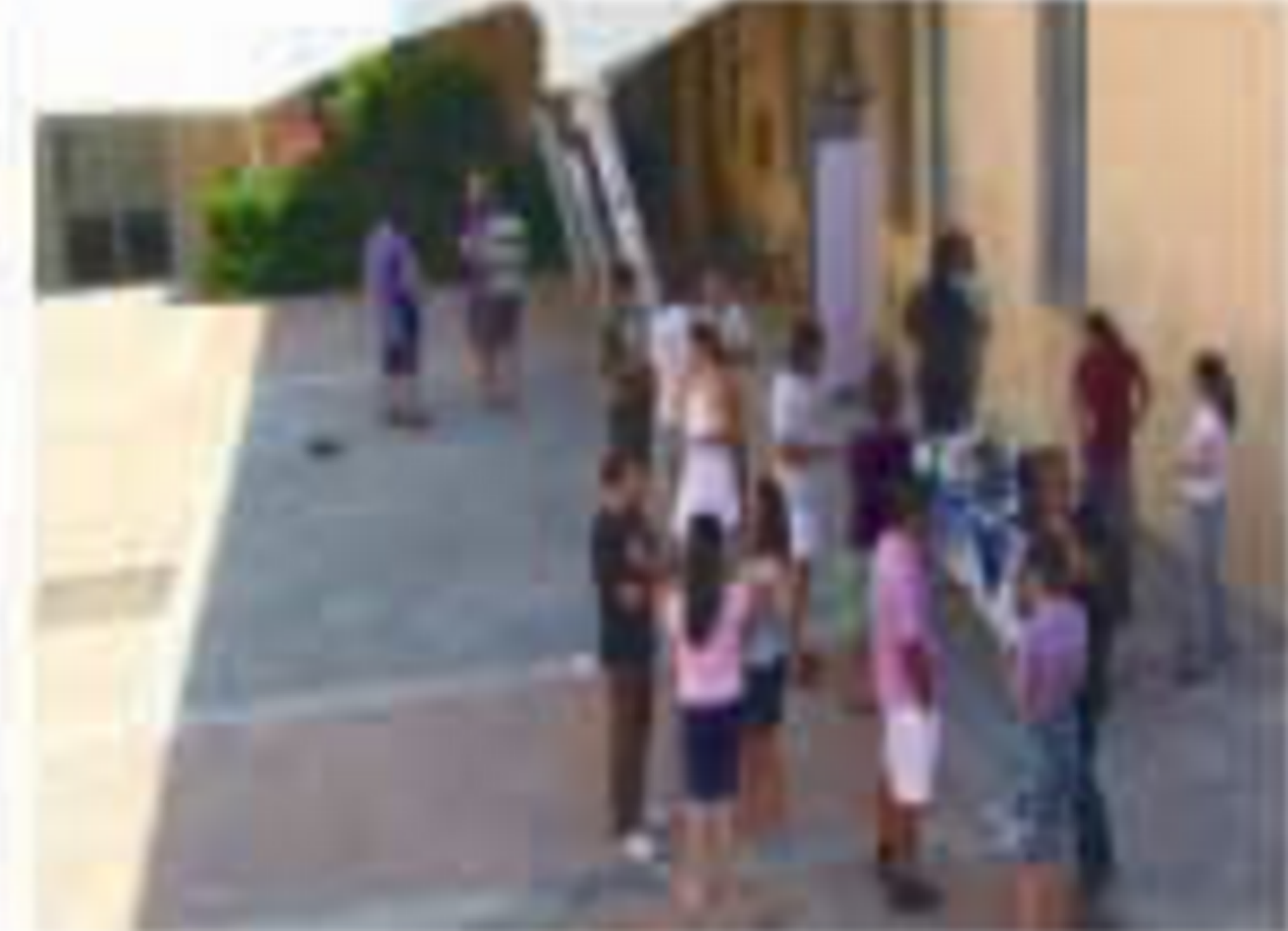}
    \includegraphics[height=27mm]{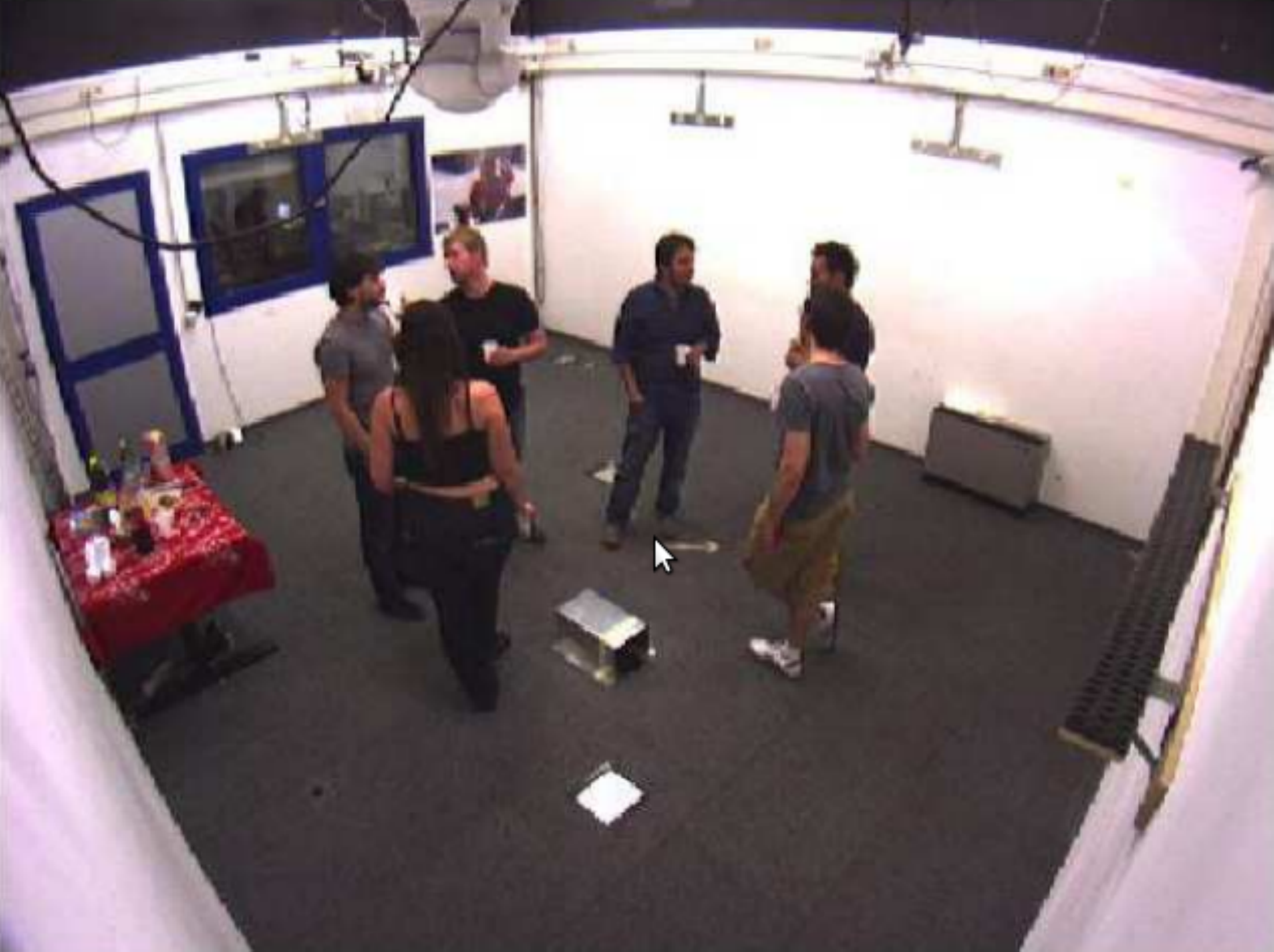}
    \caption{Top: Multiagent multigroup social intelligence at a party. Bottom: Coffee Break \cite{Cristani2011} and Cocktail Party \cite{Zen2010} dataset images.}
    \label{fig:party}
    \vspace{-5mm}
\end{figure}

Toward the ultimate goal of robotic social intelligence, we propose a computational model of social interaction management based on social perception of behavioral signals in the context of multiagent multigroup conversation. Consider a crowded social party like the ones depicted in Figure \ref{fig:party} to understand some important components of social intelligence. 
\begin{enumerate}
    \item \textbf{Group Formation and Social Signal Gating.} When many people are gathered together, it is natural for people to form smaller groups and engage in conversation. In order to participate effectively in a group conversation, each person needs to be able to pay attention to the people in the group, while suppressing information (both verbal and non-verbal) from other nearby groups. This type of social attention mechanism, which we term social signal gating, is a critical aspect of social perception and is closely tied to group identification (discussed later).
    \item \textbf{Dynamic Group Size and Influence.} At a social party, people dynamically move from group to group to start new conversations and leave old ones. The ability to adapt to {dynamic group size} and a {dynamic degree of influence} from nearby people is an important aspect of social intelligence.
    \item \textbf{Short-Term Memory.} Communication within a group is successful when there is an appropriate balance of conversational actions. At every moment, each group member decides their future conversational actions based on the past course of interaction (\emph{e.g.}, speak, respond or leave). For example, if one person has been talking very long, group members might become disinterested and disengage from the group. In other words, a social agent's {short-term memory} of past non-verbal and conversational interactions with neighbors is influential in determining its future social interactions. 
\end{enumerate}

\begin{figure}[h]
\vspace{-4mm}
  \begin{center}
    \includegraphics[width=0.4\textwidth]{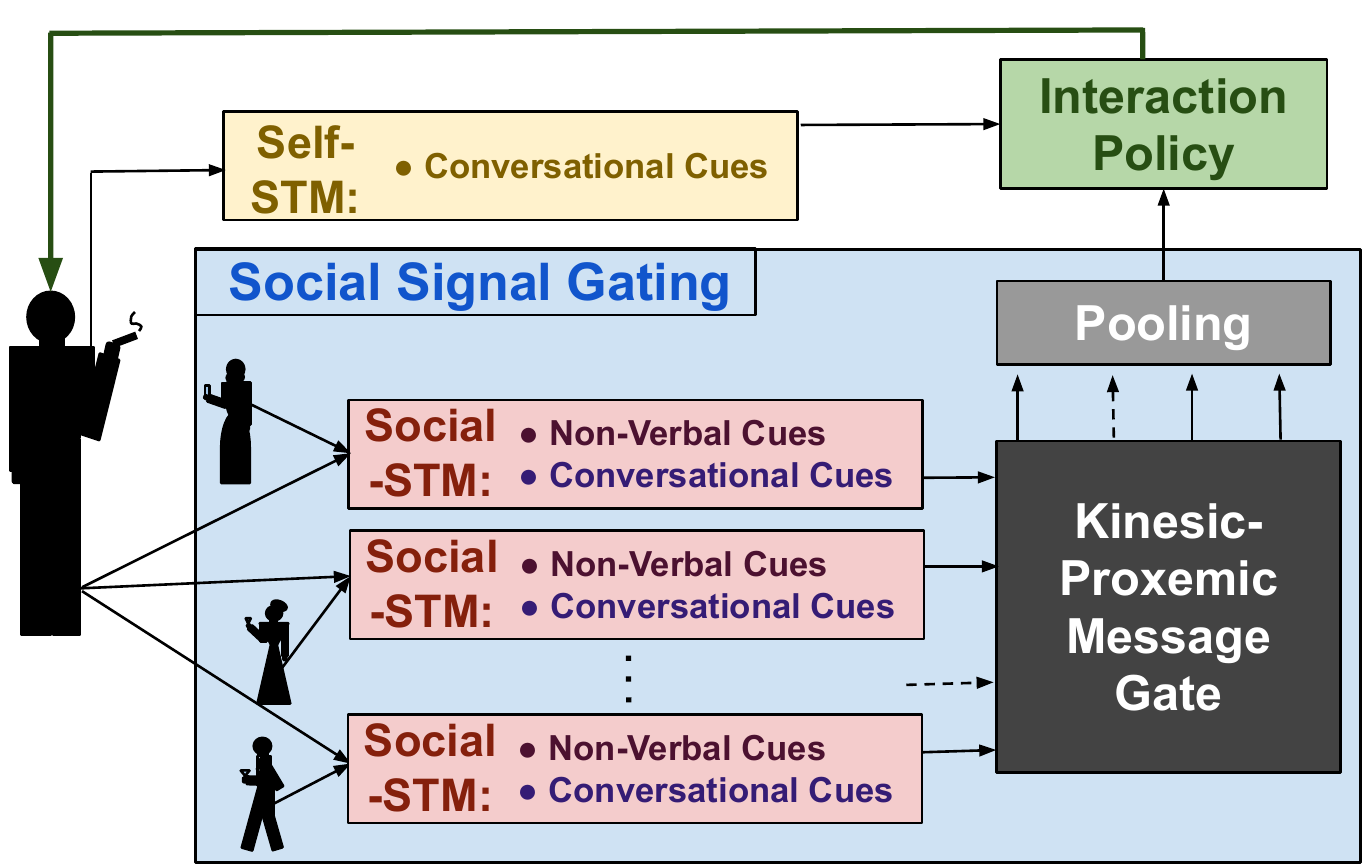}
  \end{center}
  \caption{Overview of \mgp{} Network}
  \label{fig:overview}
  \vspace{-3mm}
\end{figure}

The example above highlights critical features of a multiagent multigroup interaction model. Keeping in mind these features, we propose {\bf \mgp{}}, a \textbf{M}ultiagent \textbf{G}roup \textbf{P}erception and \textbf{I}nteraction Network. As shown in Figure \ref{fig:overview}, \mgp{} is a deep neural network consisting of the following modules: 
\begin{enumerate}
\item {\bf Social Signal Gating Module} expresses appropriate social perception by processing incoming and past social signals from neighboring agents (both non-verbal and conversational). It decides their degree of influence using a social attention function, the Kinesic-Proxemic-Message (KPM) gate. To enable reasoning with a dynamic number of influencing neighbors, it uses a pooling operation to aggregate encoded social signals. 
\item {\bf Short-term Memory (STM) Encoders} encode the agent's short-term memory of past interactions with each neighbor (Social-STM) and its own actions (Self-STM). 
\item{\bf Interaction Policy Module} processes outputs from aforementioned components to decide the agent's next conversational action.
\end{enumerate}


The \mgp{} network enables us to model social agents that make action choices in a decentralized manner. The parameters of \mgp{} are learned end-to-end via imitation learning, using a set of demonstrated group interaction sequences. In this work, demonstrations are generated by a social interaction simulator, but \mgp{} can also be learned from annotated real-world social interactions.  
Since modeling the full extent of social intelligence is very complex (\eg, power, dialogue, trust), in this work, we use abstracted expressions of social iteraction (\eg, discrete conversational actions). We believe this simplification is a necessary step towards designing more complex models of real-world social interactions.

As mentioned earlier, social perception is closely tied to group identification. Thus, we further hypothesize that, if our proposed social signal gating module is modeled correctly, we will be able to map the learned social attention directly to group membership. We demonstrate the ability of the KPM gate to identify groups in a direct, unsupervised manner, achieving state-of-the-art performance against group detection methods that use hand-defined parameters, features and complex spatial assumptions on how people arrange themselves in social situations (F-formations \cite{Kendon1990}). 

\noindent In summary, our contributions are as follows:
\begin{enumerate}
\item We present \mgp{}, a {\bf computational model} of multiagent multigroup social interactions (Section \ref{sec:architecture}).
\item In a dearth of prior work on multiagent multigroup social interaction simulators and in the absence of real-world datasets with multigroup behavioral annotations (\eg, listen, respond, strongly address), we design our own {\bf social interaction simulator}, seeking inspiration from several works which study the multi-modal nature of small-group conversational dynamics \cite{Gatica2006, katzenmaier2004identifying, gatica2005detecting, heylen2006annotating, mast2002dominance}. We use this simulated data to train and evaluate our network (Section \ref{sec:sim}).
\item We demonstrate how the ability of {\bf group identification} \emph{emerges} as a result of learning a social interaction policy, achieving competitive performance against state-of-the-art methods with the \emph{explicit} aim of detecting groups. Unlike prior work, we do so without the use of explicit group annotations, layout assumptions or hand-defined parameters (Sections \ref{sec:imitation}, \ref{sec:expt_group}). 
\end{enumerate}

\begin{figure}[h]
\vspace{-2mm}
  \includegraphics[width=0.5\textwidth]{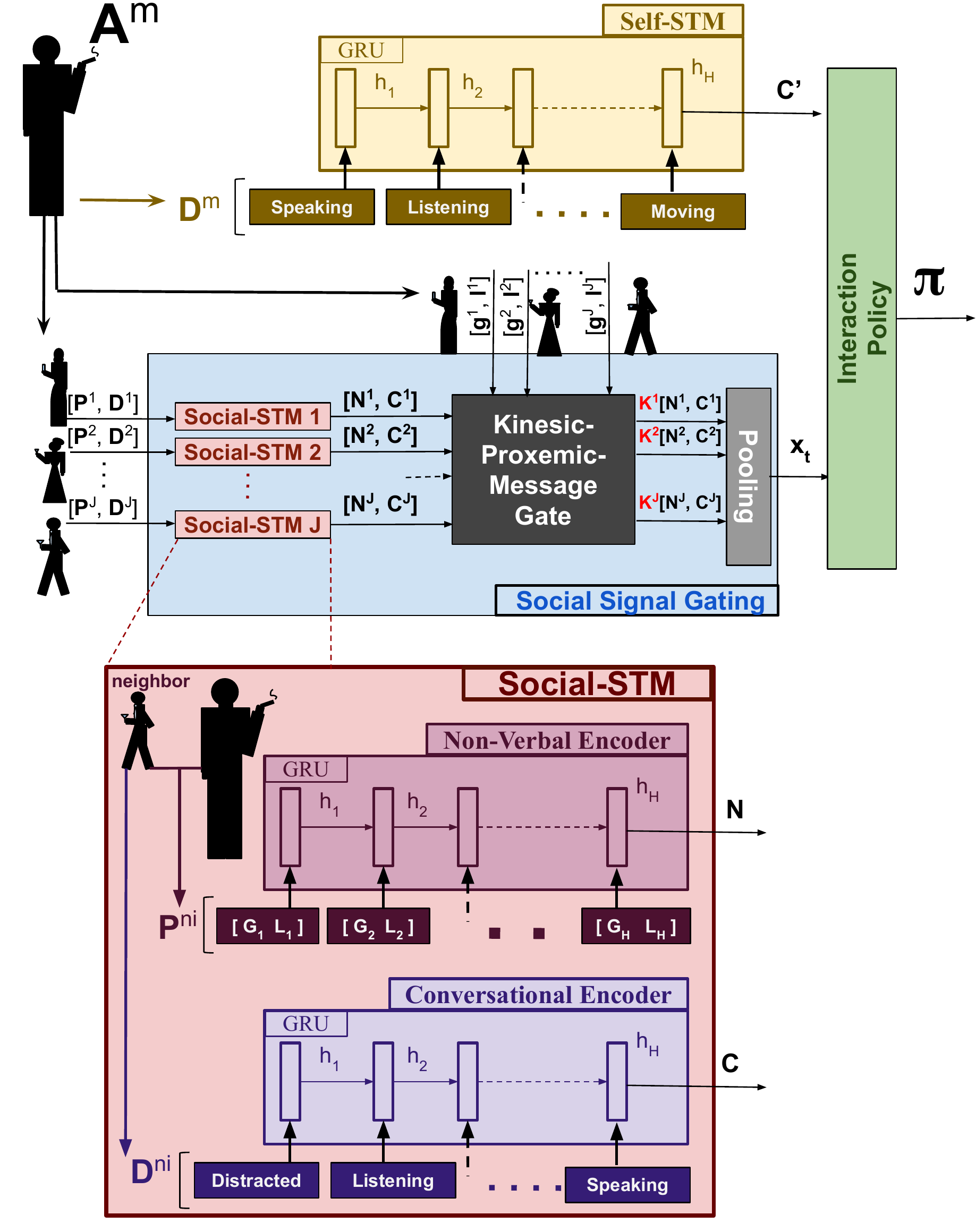}
    \caption{Components of \mgp{} Network: (1) Social Signal Gating Module, (2) Self-STM Encoder, (3) Interaction Policy. \\ A Social-STM Encoder within the Social Signal Gating Module is shown in detail. }
    \label{fig:details}
    \vspace{-3mm}
\end{figure}
  
\vspace{-3mm}
\section{Related Work} 

The analysis of intelligent agent grouping, behavior and communication is of broad interest to many disciplines including Human-Robot Interaction \cite{breazeal2003toward, breazeal2005learning, sanghvi2011automatic, castellano2012detecting, fiore2013toward, salam2016fully, yun2018automatic}, Multiagent Systems and Machine Learning \cite{Busoniu2008, Stone2000, Foerster2016, Sukhbaatar2016, Foerster2017, Le2017, Lowe2017, Peng2017, Mordatch2017, wagner2011acting, brambilla2013swarm, sanghvi2017exploiting}, Neurobiology \cite{cromwell2008sensory, freedman1987neurobiological, bronkhorst2000cocktail, shapiro1997personal, hillyard1987sensory, colino2014tactile}, Computer Vision \cite{cristani2010socially, cristani2013human, Zen2010, bazzani2013social, Setti2013, Setti2015, vascon2014game} and Psychology \cite{birdwhistell1952introduction, edward1966hall, mehrabian1967inference,davies1995mental, meltzoff2003imitation, albrecht2006social}. Since this work is focused on developing computational models of multiagent communication and identifying social groups, we elaborate on selected related work.

\vspace{-2mm}
\paragraph{\bf Human-Robot Interaction:} The effects of gaze and proxemics in human-robot interaction and social signal perception has been widely explored \cite{fiore2013toward}. Recognizing these non-verbal cues' social importance, we incorporate the perception of such cues as a central component in our model. Dialogue systems work \cite{singh2000reinforcement, serban2016building} concentrates on appropriately responding to conversational cues, while our work additionally deals with the non-verbal aspects of human communication.  Detection and automatic analysis of a human's engagement in social human-robot scenarios has been comprehensively studied \cite{sanghvi2011automatic, castellano2012detecting, salam2016fully, yun2018automatic} using various non-verbal cues such as body attitudes, facial video signals, and personality traits. Related to this, our work presents a method to automatically infer an interacting agent's attention from non-verbal and conversational cues. Sociable robots have been defined and studied in the context of their degree of interaction with humans and their functionality \cite{breazeal2003toward, dautenhahn2007socially}. Toward enabling socially intelligent robots, we present a model of multiagent multigroup interaction. The importance of simulation theory and imitative interaction in the social understanding of neighbors by robots has been studied \cite{breazeal2005learning}. We use this insight in order to learn a computational model for socially intelligent  robots through imitation of simulated interactions.

\vspace{-2mm}
\paragraph{\bf Multiagent Systems and Machine Learning:} Emergent collective behaviors have been studied in multiagent organized swarms \cite{brambilla2013swarm}. By contrast, we model individual behavior in a socially intelligent multigroup setting. Adversarial deceptive interactions have been studied between individuals \cite{wagner2011acting} and groups of intelligent robots \cite{sanghvi2017exploiting}. Instead, we assume a cooperative scenario, where interactions imitate multigroup human-to-human social reactions. Modeling multiagent communication is an active area of research in machine learning \cite{Busoniu2008,Stone2000}. Recent work makes use of deep reinforcement learning or imitation learning to discover policies for agent-to-agent communication, \eg, \cite{Foerster2016,Sukhbaatar2016,Foerster2017,Le2017,Lowe2017,Peng2017,Mordatch2017}. The motivation for these works is often collaboratively solving a particular task which might involve partial observability. Communication protocols best fit to accomplish that task are learned. By contrast, we learn to imitate social communication protocols of humans in a multigroup scenario. Some works assume the communication of internal states between agents, while others learn to communicate a particular `message' in order to solve the task. Neighbor messages might be aggregated using a pre-defined pooling operation. By contrast, we explicitly encode particular social signals between agents and learn signal gating to appropriately weight incoming signals, showing superior performance to prior pooling strategies.

\vspace{-2mm}
\paragraph{\bf Neurobiology and Signal Gating:} The concept of social signal gating introduced in our model is inspired by the neurobiological process of \emph{sensory gating} \cite{cromwell2008sensory,freedman1987neurobiological}. Sensory gating is the ability to filter out unnecessary or irrelevant external stimuli. The cocktail party effect ~\cite{bronkhorst2000cocktail} is one example of auditory sensory gating where a person is able so focus auditory attention to a specific target while ignoring other irrelevant audio input.  Similar sensory gating is observed in other senses as well, to prevent overwhelming the primary cortical areas \cite{shapiro1997personal,hillyard1987sensory, colino2014tactile}. Recognizing the importance of modelling sensory gating in socially intelligent computational systems, we use our KPM gate to mimic this phenomenon when considering the importance of nearby agents' social signals.

\vspace{-2mm}
\paragraph{\bf Social Group Interactions:} For the purpose of evaluating our model, there are no large-scale public datasets of states and conversational actions of people interacting in multiple groups. Existing datasets \cite{Cristani2011,Robicquet2016,Zen2010} only provide sequences of physical positions, gaze directions and group assignments. More importantly, no conversational action annotations (\eg, listening, speaking, \etc) are provided. Closely related to our ideas, a social simulator PsychSim \cite{marsella2004psychsim} adopts a theory-of-mind approach to modelling interactions in scenarios with agent attributes such as power and hardship. By contrast, we consider scenarios of face-to-face interactions in social situations, and require agents to have conversational roles such as responding to being addressed. Further, we aim to automatically learn the levels of reasoning required to behave appropriately in a social situation. In the absence of prior work on simulators or real-world datasets with conversational action annotations, we design our own social interaction simulator (Section \ref{sec:sim}). Inspired by several works which study the multi-modal nature of small-group conversational dynamics \cite{Gatica2006, katzenmaier2004identifying, gatica2005detecting, heylen2006annotating, mast2002dominance}, we design simulated interaction rules in order to generate data to train our network.

\vspace{-2mm}
\paragraph{\bf Social Group Identification:} There is much work on identifying the members of a group in a social situation \cite{Hung2011,Setti2013,Setti2015,Vazquez2015}, and more recently \cite{swofford2019dante}. This prior work typically assumes agents' arrangement in spatial layouts called F-formations ~\cite{Kendon1990}, often attempting to find \emph{o-space} centers \cite{Kendon1990} using heuristic strategies and hand-crafted parameters. By contrast, under no such layout assumptions, we learn parameters for a straightforward, automatic group identification mechanism (Section \ref{sec:expt_group}).

\vspace{-2mm}
\section{Social Interaction Management: The \mgp{} Network \label{sec:architecture}}

Our goal is to design a computational model for multiagent multigroup interactions that incorporates signal gating, adaptation to dynamic group sizes and short-term memory encoding. Towards this goal, we propose the \mgp{} architecture (Figures \ref{fig:overview}, \ref{fig:details}) In our design of \mgp{}, we consider several insights from social science literature: the importance of `situational awareness' and `presence' \cite{albrecht2006social}, and the impact of non-verbal social cues `kinesics' \cite{birdwhistell1952introduction} and `proxemics' \cite{edward1966hall}.
\vspace{-3mm}
\paragraph{\bf{Notation}} We denote the $m$-th agent by $A^m$ and its $J_t$ neighbors at time $t$ by $\{A^{n_i}\}^{J_t}_{i=1}$.
At time $t$, $A^m$ and its neighbor $A^{n_i}$ have gaze directions $\vct{g}_t^m, \vct{g}_t^{n_i} \in \mathbb{R}^2$, and positions $\vct{l}_t^m, \vct{l}_t^{n_i} \in \mathbb{R}^2$.  $R(\phi)$ is the rotation matrix associated with angle $\phi = \arctan{(\vct{g}_t^m)}$. Relative gaze direction and relative rotated position of $A^{n_i}$ w.r.t. $A^m$ are respectively given by:
\begin{align*} 
\vct{g}_t^{(n_i \leftarrow m)} = R(\phi)\vct{g}_t^{n_i} {\text{\;\;\;\;\;\;and\;\;\;\;\;\;}} \vct{l}_t^{(n_i \leftarrow m)} = R(\phi) (\vct{l}_t^{n_i} - \vct{l}_t^m).
\end{align*}
In our model, we use a past history of features. The history length or horizon is denoted as $H$. The histories of relative gaze directions and relative rotated positions of $A^{n_i}$ w.r.t. $A^m$ are respectively given by: 
\begin{align*}
\vct{G}_{t,hist}^{(n_i \leftarrow m)} &= \begin{bmatrix}
\vct{g}_{t-H+1}^{(n_i \leftarrow m)} & . & . & . & \vct{g}_t^{(n_i \leftarrow m)}
\end{bmatrix} \in \mathbb{R}^{2\times H},\\
\vct{L}_{t,hist}^{(n_i \leftarrow m)} &= \begin{bmatrix}
\vct{l}_{t-H+1}^{(n_i \leftarrow m)} & . & . & . & \vct{l}_t^{(n_i \leftarrow m)}
\end{bmatrix} \in \mathbb{R}^{2\times H}.
\end{align*}
The conversational action space of an agent is denoted by $\mathcal{U}$. The conversational action of $A^m$ and $A^{n_i}$ at time $t$ are represented by $|\mathcal{U}|$-dimensional one-hot vectors $\vct{d}_t^{m}$ and $\vct{d}_t^{n_i}$ respectively. The histories of conversational actions of $A^m$ and $A^{n_i}$ are respectively given by:
\begin{align*}
    \vct{D}_{t,hist}^{m} &= [\vct{d}_{t-H+1}^{m}, ..., \vct{d}_t^{m}] \in \{1,0\}^{|\mathcal{U}|\times H}\\
    \vct{D}_{t,hist}^{n_i} &= [\vct{d}_{t-H+1}^{n_i}, ..., \vct{d}_t^{n_i}] \in \{1,0\}^{|\mathcal{U}|\times H}
\end{align*}

The state of agent $A^m$ at time $t$ is given by 
\begin{align*}
    S_t^m = \left[\left\{\vct{G}_{t,hist}^{(n_i \leftarrow m)}\right\}^{J_t}_{i=1},
    \left\{\vct{L}_{t,hist}^{(n_i \leftarrow m)}\right\}^{J_t}_{i=1},
    \left\{\vct{D}_{t,hist}^{n_i}\right\}^{J_t}_{i=1},
    \vct{D}_{t,hist}^{m}\right]
\end{align*} 

\paragraph{\bf Modules} We now present details of our \mgp{} network's components. As shown in Figure \ref{fig:overview}, the \mgp{} network consists of three modules:  (a) \emph{Social Signal Gating Module:} This module aggregates observations from surrounding agents in the environment into an internal encoding; (b) \emph{Self Short-Term Memory Encoder:} This encodes each agent's own  communication action history; (c) \emph{Interaction Policy:} This module decides the agent's next action based on the agent's encoded observations of its neighbors and of itself. Specifically, the policy incorporates the outputs of the Social Signal Gating Module and Self-STM Encoder and generates a probability vector over the next action of the agent.


\subsection{Social Signal Gating Module}

Conceptually, the Social Signal Gating (SSG) Module allows the agent to select only the most relevant social signals coming from other agents. Computationally, the SSG generates an encoding for the part of available information that depends on the neighboring agents (the part that depends on itself is described later).
As shown in blue in Figure \ref{fig:details}, SSG consists of three successive computational units: (1) Social Short-Term Memory Encoder, (2) Kinesic-Proxemic-Message Gate and (3) Signal Pooling.

\paragraph{\bf Social Short-Term Memory Encoder (Social-STM)} 

For agent $A^m$, at any time $t$, the role of Social-STM is to generate an encoding for each neighbor $A^{n_i}$'s social signals. 
As shown in red in Figure \ref{fig:details}, the Social-STM contains two encoders. (1) Non-Verbal Encoder $N$ encodes a history of relative gaze directions and rotated positions $\vct{P}_{t,hist}^{n_i} = \begin{bmatrix}\vct{G}_{t,hist}^{n_i} \\ \vct{L}_{t,hist}^{n_i}\end{bmatrix} \in \mathbb{R}^{4\times H}$\; (2) Conversational Encoder $C$ encodes a history of the neighbor's conversational actions $\vec{D}_{t,hist}^{n_i}$.

Non-Verbal Encoder $N\left(\vct{P}_{t,hist}^{n_i}\right)$ summarizes $A^m$'s observations of neighbor $A^{n_i}$'s non-verbal communication history. Conversational Encoder $C\left(\vct{D}_{t,hist}^{n_i}\right)$ summarizes $A^m$'s observations of $A^{n_i}$'s past conversation.
As shown in Figure \ref{fig:details}, $N$ and $C$ are both linearly-activated gated recurrent units (GRU) \cite{Chung2014}. Each encoder learns the associated temporal dynamics and outputs neighbor `messages' $N \in \mathbb{R}^{64}$ and $C \in \mathbb{R}^{64}$. $J_t$ pairs of messages are output, one pair for each neighbor. Next, these messages are weighted by the KPM gate's importance score, pooled and passed to the interaction policy.

\paragraph{\bf Kinesic-Proxemic-Message (KPM) gate} The role of the KPM  gate is to decide how much attention to give to each neighboring agent's concatenated social signal $\left[ N\left(\vct{P}_{t,hist}^{n_i}\right), C\left(\vct{D}_{t,hist}^{n_i}\right) \right]$. The KPM gate $K\left(\vct{g}_t^{(n_i \leftarrow m)}, \vct{l}_t^{(n_i \leftarrow m)}\right)$ is a function of the relative rotated position and orientation between $A^m$ and neighbor $A^{n_i}$ at time $t$. The KPM gate is a multi-layer perceptron, consisting of two feed-forward layers, respectively activated by an exponential linear unit (ELU) \cite{Clevert2015} and a hard-sigmoid function \cite{Courbariaux2016}. For each neighbor $A^{n_i}$, it outputs a scalar `importance' weight $K \in [0,1]$, reflecting the neighbor's degree of influence on $A^m$'s next conversational action.

\paragraph{\bf Signal Pooling} The role of the signal pooling operator is to aggregate the weighted social signals for all $J_t$ neighbors of $A^m$ at time $t$. As shown in Figure \ref{fig:details}, each neighbor $A^{n_i}$'s concatenated social signal is weighted by the KPM gate's respective importance score to obtain:
\begin{align*}
    \vct{x}_t^{(n_i \leftarrow m)} = K\left(\vct{g}_t^{(n_i \leftarrow m)}, \vct{l}_t^{(n_i \leftarrow m)}\right) \;.\; \left[N\left(\vct{P}_{t,hist}^{n_i}\right) \;\; C\left(\vct{D}_{t,hist}^{n_i}\right)\right]
\end{align*}
Next, similar to \cite{Sukhbaatar2016}, we employ average pooling of weighted messages from neighbors. The pooled message $\vct{x}_t^m = \frac{1}{J_t} \sum\limits^{J_t}_{i=1} \vct{x}_t^{(n_i \leftarrow m)}$ is passed to the interaction policy $\pi$, along with the Self Short-Term Memory Encoder message (described next), to decide $A^m$'s next conversational action.

\subsection{Self Short-Term Memory Encoder (Self-STM)} 

The role of the Self-STM is to encode the agents recollection of her own past actions at time $t$. Similar to the Conversational Encoder, the Self-STM $C'$ encodes a history of the agent $A^m$'s own conversational actions $\vct{D}_{t,hist}^{m}$. As shown in yellow in Figure \ref{fig:details}, $C'(\vct{D}_{t,hist}^m)$ is a linearly-activated GRU, yielding a self `message' $C' \in \mathbb{R}^{64}$. This is concatenated with the pooled neighbor message $\vec{x}^m_t$ and passed to the interaction policy to decide $A^m$'s next conversational action.

\subsection{Interaction Policy} 

The final module of \mgp{} is the interaction policy $\pi$ which, at time $t$, decides the agent's next action based on the agent's observations of its neighbors and itself. Formally, the policy $\pi$ takes as input the concatenated outputs $\left[\vct{x}_t^m,\;\; C'\left(\vct{D}_{t,hist}^m\right)\right]$ from the SSG Module and the Self-STM (shown in green in Figures \ref{fig:overview}, \ref{fig:details}). The interaction policy $\pi\left(\left[\vct{x}_t^m,\;\; C'\left(\vct{D}_{t,hist}^m\right)\right]\right)$ is modeled as two fully-connected layers, respectively activated by an ELU and soft-max function. It yields a $|\mathcal{U}|$-dimensional output, representing a probability distribution over the agent $A^m$'s next conversational action.

\begin{figure*}[t]
\centering
\includegraphics[width=0.9\linewidth]{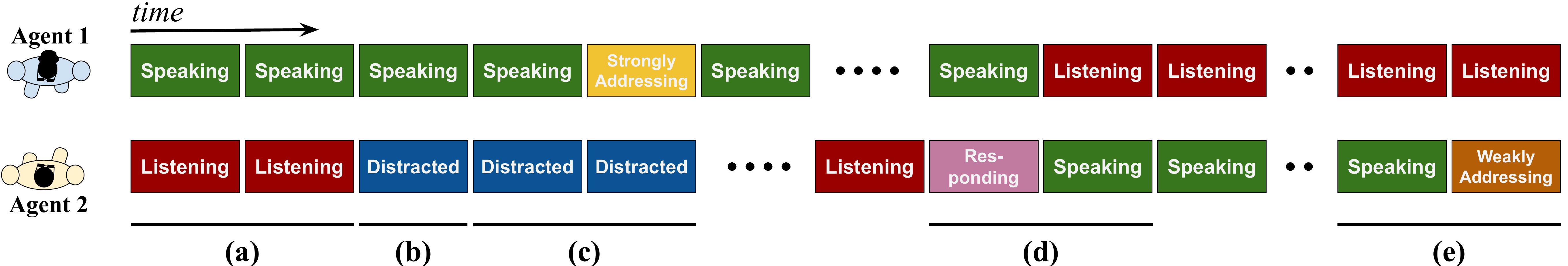}
\vspace{-2mm}
\caption{{\bf Examples of simulated communications} (static two agent case)}
\label{fig:tt_example}
\vspace{-3mm}
\end{figure*}

\section{Imitation Learning and Group Identification}
\label{sec:imitation}

By modeling the \mgp{} Network's modules $N$, $C$, $G$, $C'$, and $\pi$ as fully-differentiable functions, we ensure that it is trainable end-to-end via back-propagation, using demonstrations of multiagent multigroup state and action sequences {\footnote{Our code may be found here: \url{https://github.com/navysanghvi/MGpi}}}.

\paragraph{\bf Imitation Learning} We perform imitation learning to learn a policy from several `expert' demonstrations. In this work, we adopt the approach of behavior cloning, which enables \mgp{} to learn from a set of demonstrations $\mathcal{D}$ directly, via supervised learning without needing to explicitly learn state transitions or reward functions. Similar approaches can be found in recent work on multiagent behavior prediction~\cite{Alahi2016,Robicquet2016}. We explain how these demonstrations are obtained in Section \ref{sec:sim}.
Each demonstration $D \in \mathcal{D}$ is of length $T$, and denoted by $D=\left\{(\vct{S}_{t}, \vct{U}_{t})\right\}^{T}_{t=1}$, where $\vct{S}_t$ and $\vct{U}_t$ are the joint state and joint action of the $M$ interacting agents at time $t$. Therefore, $\vct{S}_t = \left\{\vct{S}^m_t\right\}^M_{m=1}$, and $\vct{U}_t = \left\{\vct{d}^m_t\right\}^M_{m=1}$, where $\vct{S}^m_t$ and $\vct{d}^m_t$ are individual agent state and action, as described in the notation in Section \ref{sec:architecture}. Note that $\vct{S}_{t}$ contains no information about the group each agent belongs to.

\paragraph{\bf Group Assignments} We emphasize again that training of the \mgp{} Network requires {no supervision with respect to group assignments}. Moreover, we identify these assignments implicitly using the KPM gate module $K$. For every agent $A^m$, in order to minimize the error between \mgp{}'s predicted action $\hat{\vct{d}}^m_{t}$ and demonstrated action $\vct{d}^m_{t}$, KPM gate $K$ must implicitly learn spatial attention. The higher the `importance' weight assigned by $K$ to a neighbor's encoded non-verbal and conversational history, the more likely the neighbor is to be in the same group as the $A^m$, and the greater its influence on $A^m$'s resulting policy $\pi$. 

\paragraph{\bf Group Identification} Once the \mgp{} network parameters are learned, we can use the KPM gate activations to {identify groups}. In order to identify groups, at each time step, we define a distance $D(n,j)$ between two agents $n,j$ based on the output of learned gating function $K$: $D(n,j) = 1 - \frac{1}{2}(K(\vct{g}^{(j \leftarrow n)}, \vct{l}^{(j \leftarrow n)}) + K(\vct{g}^{(n \leftarrow j)}, \vct{l}^{(n \leftarrow j)}))$. The distance is simply computed from the weighted average of the bi-directional weights computed by the KPM gate. We can compose an affinity matrix of pairwise distances between all agents and run the DBSCAN clustering algorithm~\cite{Ester1996} to cluster people into conversational groups. As mentioned in our Introduction, this is a relatively simple and straight-forward approach. It is compared in Section \ref{sec:expt_group} with state-of-the-art approaches based on complex \emph{o-space} estimation under assumptions of particular spatial layouts called F-formations \cite{Kendon1990}.
\begin{table*}[t]
\caption{Results for Static Scenario: Action Prediction (mean average precision), and Test Loss (cross-entropy)}
\vspace{-3mm}
\centering
\scalebox{1}{
\begin{tabular}{lcccc}
\toprule
      Model {\bf (mAP's)} &  $J=2$& 4&8&12 \\ 
     \midrule
     Neighbor States Only \hfill (\texttt{NSO}) & 0.67 & 0.69 & 0.70 & 0.68 \\
     Self State Only \hfill (\texttt{SSO}) & 0.76 & 0.76 & 0.76 & 0.76 \\
     Equal Pooling \hfill (\texttt{EQPOOL}) &0.87 & 0.86& 0.85 & 0.85\\
     Social Pooling \hfill (\texttt{SOCPOOL}) &0.85 & 0.84 & 0.84 & 0.84\\
     \midrule
     {\bf \mgp{} Network} & {\bf 0.88} & {\bf 0.88} & {\bf 0.89} & {\bf 0.88}\\
\bottomrule
\end{tabular}
}
\hspace{4mm}
\scalebox{1}{
\begin{tabular}{lcccc}
\toprule
      Model {\bf (cross-entropy losses)} &  $J=2$& 4&8&12 \\ 
     \midrule
     Neighbor States Only \hfill (\texttt{NSO}) & 0.99 & 0.95 & 0.89 & 1.00 \\
     Self State Only \hfill (\texttt{SSO}) & 0.29 & 0.29 & 0.29 & 0.29 \\
     Equal Pooling \hfill (\texttt{EQPOOL}) &0.21 & 0.22 & 0.22 & 0.22\\
     Social Pooling \hfill (\texttt{SOCPOOL}) &0.21 & 0.22 & 0.22 & 0.25\\
     \midrule
     {\bf \mgp{} Network} & {\bf 0.20} & {\bf 0.21} & {\bf 0.20} & {\bf 0.21}\\
\bottomrule
\end{tabular}
}
\label{tab:action_prediction1}
\end{table*}

\begin{table*}[t]
\caption{Results for Dynamic Scenario: Action Prediction (mean average precision), and Test Loss (cross-entropy)}
\vspace{-3mm}
\centering
\scalebox{1}{
\begin{tabular}{lcccc}
\toprule
      Model {\bf (mAP's)} &  $J=2$& 4&8&12 \\ 
     \midrule
     Neighbor States Only \hfill (\texttt{NSO}) & 0.57 & 0.62 & 0.65 & 0.64 \\
     Self State Only \hfill (\texttt{SSO}) & 0.78 & 0.78 & 0.78 & 0.78 \\
     Equal Pooling \hfill (\texttt{EQPOOL}) &0.85 & 0.86 & 0.85 & 0.85\\
     Social Pooling \hfill (\texttt{SOCPOOL}) &0.85 & 0.85 & 0.86 & 0.85\\
     \midrule
     {\bf \mgp{} Network} & {\bf 0.87} & {\bf 0.87} & {\bf 0.88} & {\bf 0.88}\\
\bottomrule
\end{tabular}
}
\hspace{4mm}
\scalebox{1}{
\begin{tabular}{lcccc}
\toprule
      Model {\bf (cross-entropy losses)} &  $J=2$& 4&8&12 \\ 
     \midrule
     Neighbor States Only \hfill (\texttt{NSO}) & 1.24 & 1.12 & 1.05 & 1.07 \\
     Self State Only \hfill (\texttt{SSO}) & 0.31 & 0.31 & 0.31 & 0.31 \\
     Equal Pooling \hfill (\texttt{EQPOOL}) &0.26 & 0.26 & 0.25 & 0.25\\
     Social Pooling \hfill (\texttt{SOCPOOL}) &0.27 & 0.27 & 0.25 & 0.26\\
     \midrule
     {\bf \mgp{} Network} & {\bf 0.25} & {\bf 0.23} & {\bf 0.25} & {\bf 0.22}\\
\bottomrule
\end{tabular}
}
\label{tab:action_prediction2}
\vspace{-2mm}
\end{table*}

\section{Experiments: Social Interaction Simulator}
\label{sec:sim}

\paragraph{\bf Conversational Action Data} 
We aim to learn computational models that can encode the kinesics and proxemics in group formation and conversational actions (\eg, responding, distraction, \etc) in multiagent multigroup scenarios. Unfortunately, there are no public datasets of large-scale demonstrations of the physical positions, gaze directions, group assignments, and communication actions of multiple socially interacting people. Existing datasets \cite{Cristani2011,Robicquet2016,Zen2010} only provide sequences of the physical positions, gaze directions and group assignments (but annotated only sparsely). More importantly, \emph{no conversational action annotations are provided}. Similarly, there are no social interaction simulators for the kind of face-to-face behavioral actions of interest to us. In order to evaluate the capacity of our proposed model, we simulate multigroup human communication, drawing inspiration from several prior studies and observations of small-group conversational dynamics \cite{Gatica2006, katzenmaier2004identifying, gatica2005detecting, heylen2006annotating, mast2002dominance}. While we simulate sequences of non-verbal interactions, we point out that the physical layouts of agents are from real-world datasets.

\paragraph{\bf Initial Agent Layout Data:} We experiment with both artificially constructed and real-world data to form our initial multigroup layouts - specifically, we use (a) Synthetic ~\cite{Cristani2011}\footnote{\url{http://profs.sci.univr.it/~cristanm/ssp/}}, (b) Coffee Break ~\cite{Cristani2011}, and (c) Cocktail Party \cite{Zen2010} datasets to provide 100, 119, and 320 non-identical, independent layouts respectively. Multiple people (6-12 per layout) are organized into several groups (Fig. \ref{fig:party}).

\begin{figure}[h]
\centering
    \begin{subfigure}[]{0.22\textwidth}
    \centering
    \includegraphics[width=\columnwidth]{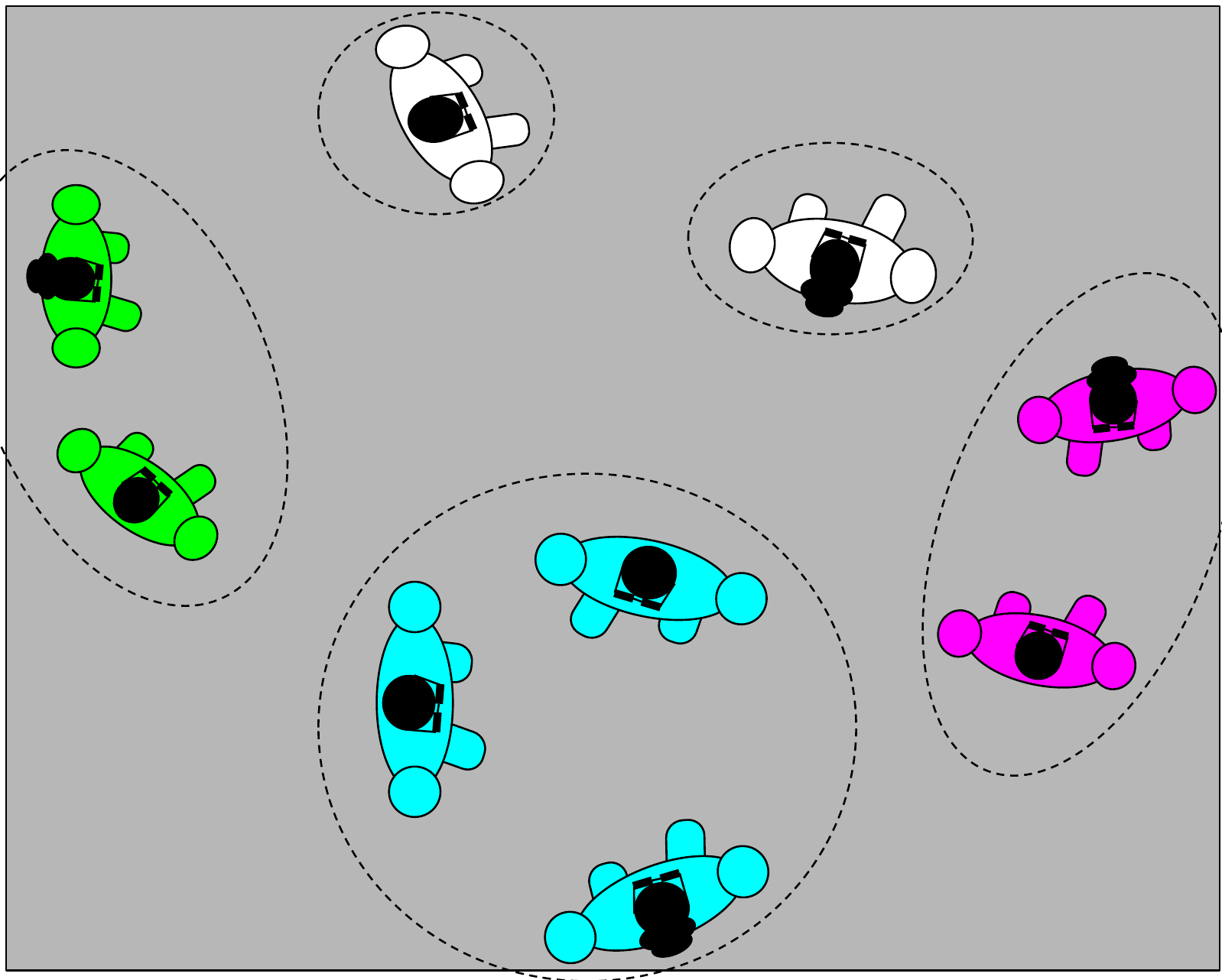}
    \end{subfigure}
    \centering
    \begin{subfigure}[]{0.22\textwidth}
    \centering
    \includegraphics[width=\columnwidth]{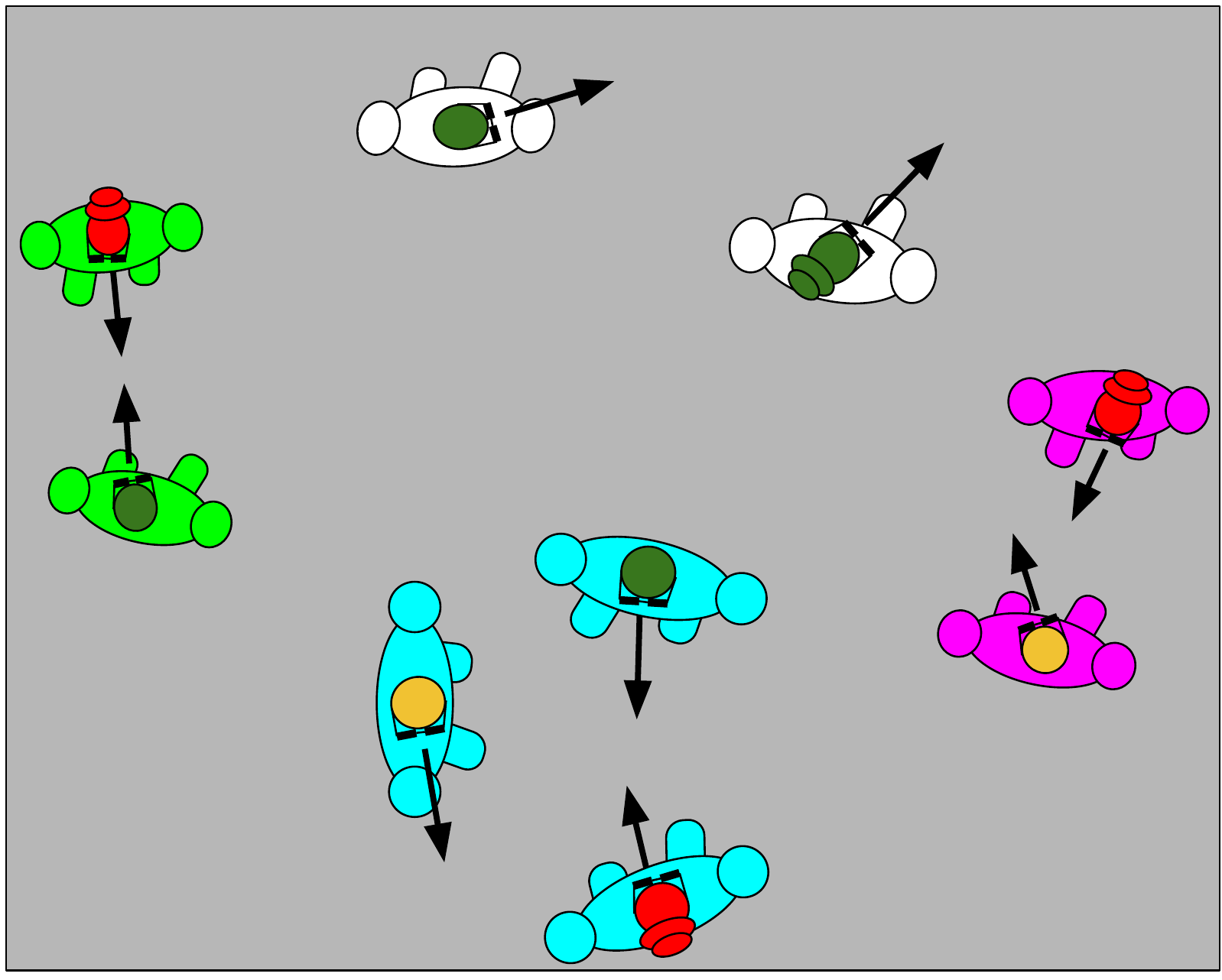}
    \end{subfigure}
\caption{{\bf Example physical layout} \cite{Cristani2011} and subsequent conversational interaction. Arrows represent gaze directions, head colors indicate conversational action - red: speaking, yellow: distracted, green: listening. }
\label{fig:synth}
\vspace{-4mm}
\end{figure}

\paragraph{\bf Rules of Interaction:} 
To design multigroup communication protocols, we draw inspiration from several observations of small-group conversational dynamics in literature, namely: speaking and turn-taking,  \cite{Gatica2006}, addressing \cite{katzenmaier2004identifying}, interest in meetings \cite{gatica2005detecting, heylen2006annotating}, and the relation of speaking time and dominance \cite{mast2002dominance}. 
We design per-group rules for two scenarios: {\bf (1) Static scenario}, where agents stay within a social group and do not transition to other groups. Six conversational actions are considered $\mathcal{U}^{stat} = \{$\textit{Speaking}, \textit{Listening}, \textit{Distracted}, \textit{Strongly Addressing}, \textit{Weakly Addressing}, \textit{Responding}$\}$.  {\bf (2) Dynamic scenario}, where agents transition from group to group, leading to a temporally evolving group assignments. Seven conversational actions are considered $\mathcal{U}^{dyn} = \mathcal{U}^{stat} \cup \{$\textit{Moving}$\}$. Based on observations from literature described above, we design per-group probabilistic rules conditioned on a temporal history of past group actions, to decide the evolution of each agent's conversational actions. The same rules are applied to every group in the scene. We simulate 600-step interaction episodes starting from each set of initial layouts ((a),(b) and (c) above) to obtain training data. Figure~\ref{fig:synth} shows an example of initial multigroup layout and subsequent interaction{\footnote{Example simulator videos may be found here: \url{https://github.com/navysanghvi/MGpi}}}.

We reiterate: group assignment information is used \emph{only for evaluation} and \emph{not} for supervisory training signals to \mgp{}. Instead, \mgp{} implicitly identifies groups using the KPM gate.  We hypothesize that our automatic KPM gating mechanism will learn to discover relevant neighbors in both static and dynamic scenarios. While we recognize the gap between reality and our simulations, we believe that, in the absence of existing data or other simulators, this is a necessary first step towards understanding how to model multiagent multigroup interaction.

\paragraph{\bf Two-Agent Simulation Example} Using Figure \ref{fig:tt_example}, we illustrate our rules in a simple two-agent case. In the static scenario, as shown in Figure \ref{fig:tt_example}(a), one agent is \textit{Speaking} while the other is \textit{Listening}. The \textit{Listening} agent may become \textit{Distracted} (Figure \ref{fig:tt_example}(b)) with a certain probability. After some time steps, the \textit{Speaking} agent may start \textit{Strongly Addressing} the group, to draw back the attention of the \textit{Distracted} member (Figure  \ref{fig:tt_example}(c)). Once the \textit{Speaking} agent has spoken for some time, it yields to the other. The other agent is then \textit{Responding} followed by \textit{Speaking} (Figure \ref{fig:tt_example}(d)). The \textit{Speaking} agent might transition to \textit{Weakly Addressing} (Figure \ref{fig:tt_example}(e)). Additionally, in a dynamic scenario, a group member who is \textit{Distracted} may start \textit{Moving} toward another group. When a \textit{Moving} agent joins a new group, it is welcomed as the next speaker.

\paragraph{\bf Baselines}  We now describe experiments and results of applying the \mgp{} network to predict the communication actions of simulated social agents. We compare the \mgp{} network with the following baselines, including communication models used in prior work:

\textbf{{Neighbor State Only (\texttt{NSO}):}} We omit Self-STM Encoder $C'$ so as to only use cues from neighbors. This helps us test whether observations of the self are necessary for modeling social interaction.

\textbf{{Self State Only (\texttt{SSO}):}} We omit the Social Signal Gating Module so as to use cues of the agent's self alone.  This helps us test whether observations of nearby agents' social signals are necessary for modeling social interaction.

\textbf{{Equal Pooling (\texttt{EQPOOL}):}} Prior work studies the role of message broadcasting in scenarios where one agent simply averages messages it receives from other agents \cite{Foerster2016,Mordatch2017,Sukhbaatar2016}. We implement this message broadcasting baseline by omitting the KPM gate $K$ so that all neighbor encoded `messages' $N$ and $C$ are \textit{equally weighted}, such that $\vct{x}_t^{(n_i \leftarrow m)} = \left[N\left(\vct{P}_{t,hist}^{n_i}\right) \;\; C\left(\vct{D}_{t,hist}^{n_i}\right)\right]$.  This helps us test if equal social perception to neighbor agents is sufficient for modeling social interaction.

\textbf{{Social Pooling (\texttt{SOCPOOL}):}} We adopt social pooling \cite{Alahi2016,Lee2017} as a second message broadcasting baseline. It uses a grid that divides the 2D world into non-overlapping regions and pools messages by region. In our experiments, we use a $4\times 4$ grid around target locations, each of size $50\times 50$ pixels (this configuration yielded best performance). For each region, we average $\vct{x}_t^{(n_i \leftarrow m)}$ over neighbors $A^{n_i}$ located in that region. For this baseline, the KPM gate and original Signal Pooling Mechanism of \mgp{} are replaced by an operator that performs grid-wise pooling and concatenates pooled messages. We expect this baseline to learn to implicitly gate messages by down-weighting messages in far away grids.


\paragraph{\bf Training/Evaluation Scheme} Demonstrations of multigroup communication were collected as described previously. For each layout in each dataset (Synthetic, Coffee Break and Cocktail Party), agents changed modes 600 times, \ie, $T=600$. We split demonstrations into two subsets with equal number of layouts from each dataset, and learned models on one subset to test the other. All metrics are averaged over the two test subsets for two-fold cross-validation. All models were trained for 30 epochs with mini-batches of size 4096, various numbers of agents $J \in\{2,4,8,12\}$, and a history window $H = 15$, and we confirmed that neither further training nor smaller mini-batches improved performance. Each network was trained to minimize categorical cross entropy (also called negative log likelihood) between predicted probabilities of actions $\{\{\hat{\vct{U}}_{t}\}^T_{t=1}\}_{\mathcal{D}}$ and actual demonstrated actions  $\{\{\vct{U}_{t}\}^T_{t=1}\}_{\mathcal{D}}$ via Adam \cite{Diederik2014}. 
We report the cross-entropy loss on testing subsets. As every model assigns maximum probability to one of six (static scenario) or seven (dynamic scenario) actions, we also evaluate the model on mean average precision (mAP) over actions.


\begin{table*}[htbp!]
\caption{{ Evaluation on group detection}: Precision, Recall and F1 scores under the $|G|$ criterion. \\State-of-the-art methods DS~\cite{Hung2011}, HVFF-ms~\cite{Setti2013}, GCFF~\cite{Setti2015}, and GRUPO~\cite{Vazquez2015} are compared against a simple baseline and the KPM gate (ours). Standard deviation across 5 training runs are shown in brackets. Due to averaging over a lesser number of datasets, italicized results for DS and GRUPO cannot be taken into account for a fair comparison}
\vspace{-2mm}
\centering
\scalebox{1}{
\begin{tabular}{c|c|cccc||c|cc}
\toprule
     \multicolumn{2}{c|}{Method$\Rightarrow$} & \multirow{2}{*}{DS} & \multirow{2}{*}{HVFF-ms} &  \multirow{2}{*}{GCFF} & \multirow{2}{*} {GRUPO} & {Pose-only} & \multicolumn{2}{c}{\bf KPM gate (ours)}\\\cline{1-2}\cline{8-9}
     {Dataset $\Downarrow$} & {Metric $\Downarrow$} & & & & & {Baseline} & {\bf Static Scenario}&{\bf Dynamic Scenario} \\
     \midrule
     \multirow{3}{*}{Synthetic} & Precision & {0.68} & {0.72} & {0.91} & {N/A} & 0.88 & {\bf 1.00} {\footnotesize{(0.000)}}  & {\bf 1.00} {\footnotesize{(0.000)}}\\
     & Recall & 0.80 & 0.73 & 0.91 & N/A & 0.58 & {\bf 0.96} {\footnotesize{(0.008)}}& {\bf 0.98} {\footnotesize{(0.000)}} \\
     & F1 Score & {0.74} & {0.73} & {0.91} & {N/A} & 0.70 & {\bf 0.98} {\footnotesize{(0.004)}}  & {\bf 0.99} {\footnotesize{(0.000)}} \\
     \midrule
     \multirow{3}{*}{Coffee Break} & Precision  & 0.40 & 0.40 & 0.61 & N/A & 0.53 & {\bf 0.63} {\footnotesize{(0.004)}} & {\bf 0.61} {\footnotesize{(0.004)}}\\
     & Recall & 0.38 & 0.38 & {\bf 0.64} & N/A & 0.46 & {0.63} {\footnotesize{(0.010)}} & {0.62} {\footnotesize{(0.005)}} \\
     & F1 Score &  0.39 & 0.39 & {\bf 0.63} & N/A & 0.49 & {\bf 0.63} {\footnotesize{(0.005)}} & {0.62} {\footnotesize{(0.001)}} \\
     \midrule
     \multirow{3}{*}{Cocktail Party} & Precision & N/A & 0.30 & {0.63} & {\bf 0.65} & 0.29 & 0.60 {\footnotesize{(0.004)}} & {0.63} {\footnotesize{(0.010)}}\\
     & Recall & N/A & 0.30 & {\bf 0.65} & 0.63 & 0.27 & 0.56 {\footnotesize{(0.007)}} & 0.55 {\footnotesize{(0.003)}} \\
     & F1 Score & N/A & 0.30 & {\bf 0.64} & {\bf 0.64} & 0.28 & 0.58 {\footnotesize{(0.005)}} & 0.59 {\footnotesize{(0.005)}} \\
     \midrule
     \midrule
     \multirow{3}{*}{ Overall Mean} & { Precision} & {\emph{0.54}} & 0.47 & 0.71 & {\emph{0.65}} & 0.57 & {\bf 0.74} {\footnotesize{(0.002)}} & {\bf 0.75} {\footnotesize{(0.004)}}\\
     & { Recall} & {\emph{0.59}} & 0.47 & {\bf 0.73} & {\emph{0.63}} & 0.44 & 0.72 {\footnotesize{(0.008)}} & 0.72 {\footnotesize{(0.002)}} \\
     & { F1 Score} & {\emph{0.56}} & 0.47 & 0.72 & {\emph{0.64}} & 0.49 & {\bf 0.73} {\footnotesize{(0.004)}} & {\bf 0.73} {\footnotesize{(0.002)}} \\
    \bottomrule
\end{tabular}
}
\label{tab:group}
\vspace{-3mm}
\end{table*}

\paragraph{\bf Results} Tables \ref{tab:action_prediction1} and \ref{tab:action_prediction2} shows all model performances in communication action prediction, with $H=15$, and various numbers of neighbors $J$. We emphasize that $J$ is only fixed during training; since a single Social-STM Encoder is learned, \emph{any} number of neighbors may be considered at every time step during testing - in fact, \emph{different} numbers of neighbors may be considered for each agent in the scene. 

As hypothesized, \mgp{} effectively learns protocols in both static and dynamic scenarios, outperforming strong baselines in terms of both mean average precision (mAP) and test loss, validating our choices of various encoding modules and demonstrating each one's necessity. Confusion matrices of action prediction for all networks 
(omitted here for spcae) show that \texttt{NSO} and \texttt{SSO} work complementarily: \texttt{NSO} predicts better if agents make strong or weak addressing decisions by observing others. On the other hand, \texttt{SSO} performs better when observation of the self is necessary, \ie, in order to predict if agents keep speaking, listening, are distracted or responding. \mgp{} outperforms baselines for most actions, both in the static and dynamic case, with lower ambiguity. 

\begin{figure}[h]
\vspace{-1mm}
\centering
\includegraphics[width=0.25\linewidth]{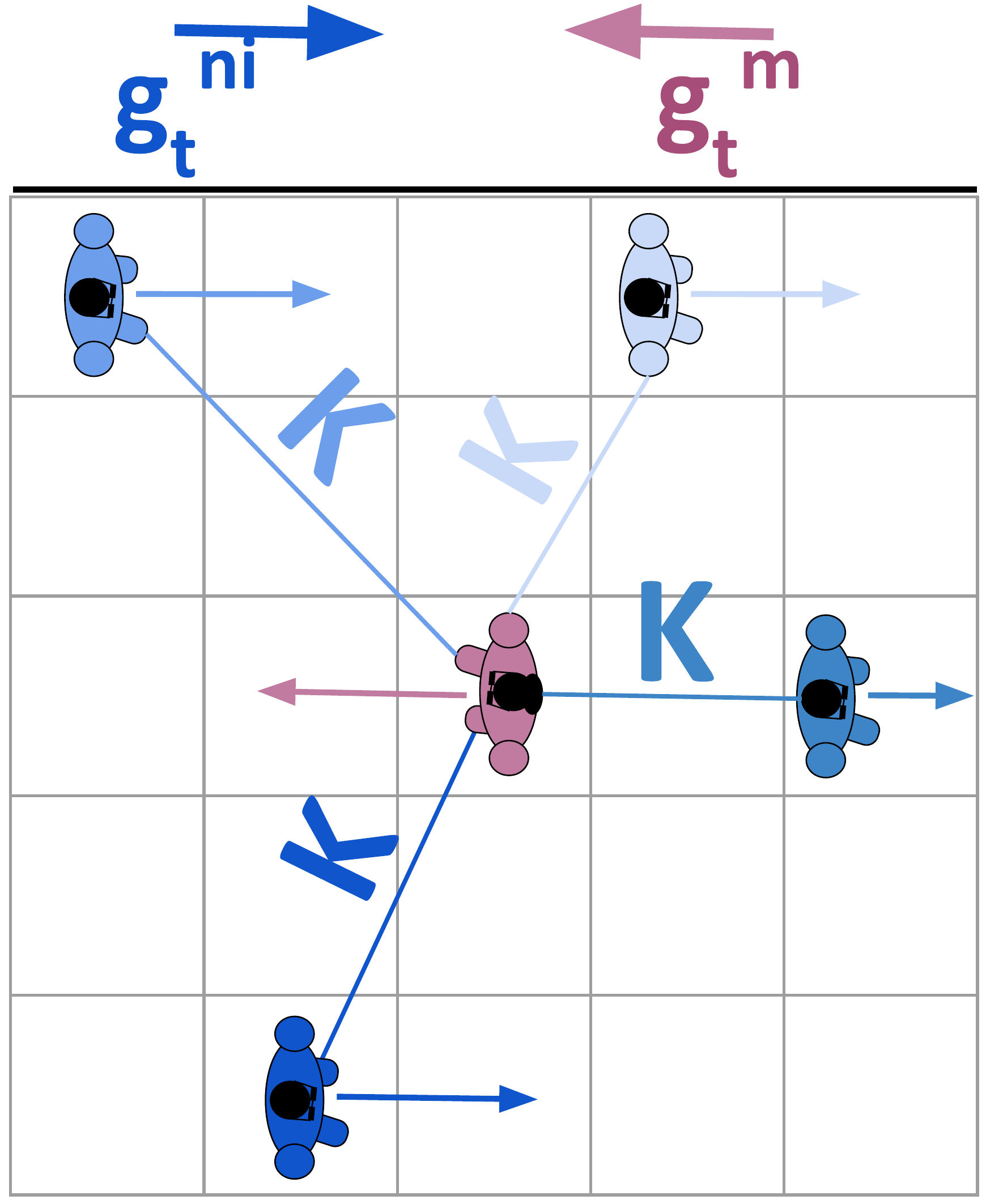}
\includegraphics[width=0.74\linewidth]{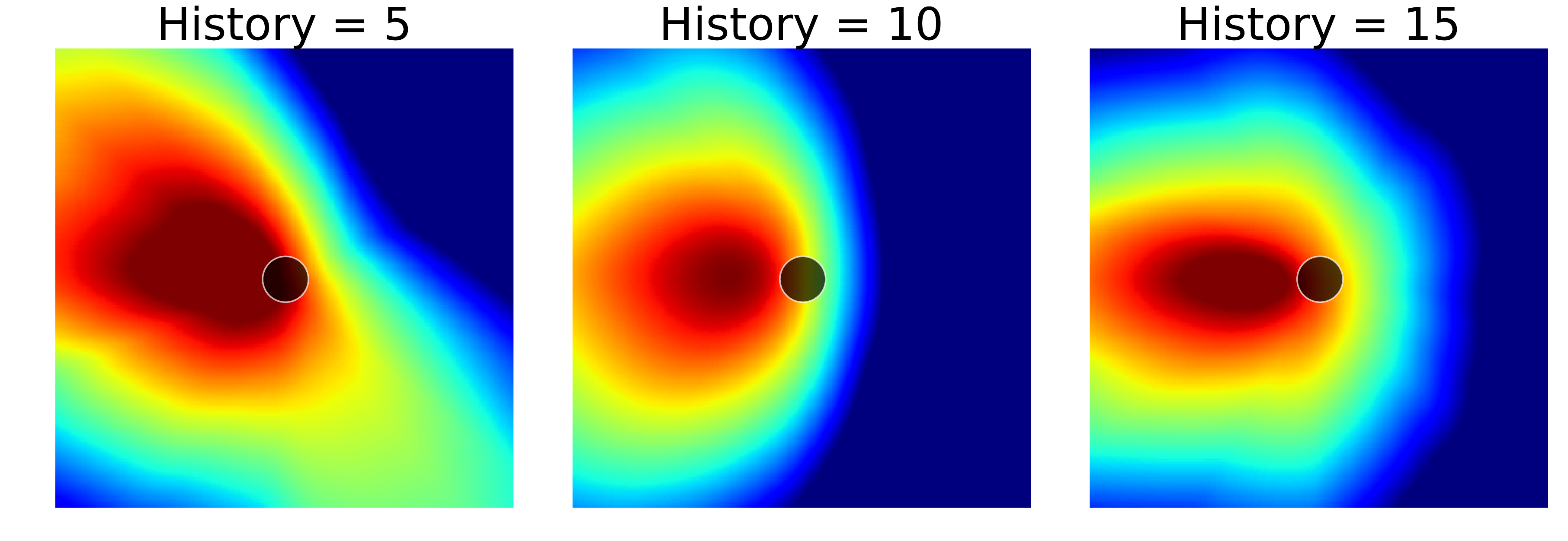}
\caption{{\bf Social Attention of KPM Gate:} For various positions of a neighbor $A^{n_i}$ (\eg, any blue figure) in a fine grid around $A^m$ (magenta figure, black circles), warmer colors indicate higher outputs from the learned KPM gate $K$. Gaze directions $\vct{g}^{m}_t = [-1,0]$,\;\; $\vct{g}^{n_i}_t = [+1,0]$.\;\; History windows $H \in \{5, 10, 15\}$.}
\label{fig:visualization1}
\vspace{-3mm}
\end{figure}

\paragraph{\bf Learned Social Attention} Figure~\ref{fig:visualization1} represents physical constraints learned by the \mgp{} Network's KPM gate $K$ to judge if neighbor agent $A^{n_i}$ is in the same group as agent $A^m$. These visuals are obtained after training in the static scenario. We visualize the output from $K$ at various locations around $A^m$ while fixing relative gaze $g^{(n_i \leftarrow m)}$. As expected, we see improved learning with larger history windows $H$, and choose $H = 15$ in our experiments. We confirm that increasing the history window does not improve performance, suggesting that, once trained on real-world scenarios, \mgp{} might be able to \emph{reflect the optimal length of short-term memory exhibited in humans}. Also as expected, we see a higher output from our KPM gate when agents are looking toward each other, i.e., when the neighbor $A^{n_i}$ is toward the left of the image, looking in $+x$ direction, \emph{i.e.}, toward $A^m$, who is looking in $-x$ direction.

\section{Experiments: Group Identification in Real Data}\label{sec:expt_group}

We have demonstrated the effectiveness of the \mgp{} network in learning group communication policy with simulated social agents. We now test the effectiveness of our learned automatic social signal gating mechanism, the KPM gate, in identifying communication groups in real-world data ~\cite{Cristani2011,Zen2010}. 

\paragraph{\bf Training \mgp{}} Similar to Section \ref{sec:sim}, we train \mgp{} using a dataset of simulated interactions from a mixture of initial layouts from the Synthetic~\cite{Cristani2011}, Coffee Break~\cite{Cristani2011} and Cocktail Party~\cite{Zen2010} datasets. For the sake of equal representation in our training data, we use 100 initial layouts from each dataset, for a total of 300 demonstrations of $T=600$ steps each. Each dataset has several continuous frames with various numbers of people annotated with positions, orientations, and group assignments (Fig. \ref{fig:party}). \mgp{} is trained for 20 epochs, in both static and dynamic scenarios, setting $J = 12$, which is sufficiently larger than group membership (6 at most) in the datasets. This ensures that the network must learn to filter out irrelevant social cues from non-group agents, strengthening the contribution of the KPM gate to good network performance. Further, this enhances the chances that KPM gate's learned social attention is effective for group identification.

\paragraph{\bf Unsupervised Group Identification} As described in Section \ref{sec:imitation}, once \mgp{} is trained, we define a distance $D(n,j)$ between two agents $n,j$ based on the output of learned gating function $K$: $D(n,j) = 1 - \frac{1}{2}(K(\vct{g}^{(j \leftarrow n)}, \vct{l}^{(j \leftarrow n)}) + K(\vct{g}^{(n \leftarrow j)}, \vct{l}^{(n \leftarrow j)}))$. For every initial layout in real-world datasets Synthetic, Coffee Break, and Cocktail Party \cite{Cristani2011,Zen2010}, we use this simply computed distance measure to form an affinity matrix of pairwise distances between agents in the scene. Using this affinity matrix, we run the DBSCAN clustering algorithm~\cite{Ester1996} to cluster people into conversational groups.

\paragraph{\bf Evaluation} Group detection performance is measured by the F1 score (harmonic mean of Precision and Recall) under the $|G|$ condition~\cite{Cristani2011,Setti2013}. Per frame, a group is judged to be detected correctly if {\emph{all}} constituent people are grouped into a single cluster. Precision, Recall and F1 scores are averaged over frames for each dataset (Synthetic, Coffee Break, Cocktail Party) separately, as well as over all datasets.
Due to the stochastic nature of training \mgp{}, we show average and standard deviation of KPM gate's performance over 5 training runs. 

\paragraph{\bf Comparisons} We define a pose-only baseline, which runs DBSCAN using actual euclidean distances between agents in each scene, scaled by the maximum such distance. Unlike \mgp{} and other state-of-the-art methods, this does not use gaze or body orientation information. In addition to this simple baseline, several state-of-the-art group identification methods DS\cite{Hung2011}, HVFF-ms \cite{Setti2013}, GCFF \cite{Setti2015}, GRUPO \cite{Vazquez2015} are chosen for comparison. These methods are based on complex \emph{o-space} estimation under assumptions of particular spatial layouts called F-formations \cite{Kendon1990}, often using heuristic parameters. By contrast, KPM gate offers a direct, unsupervised approach to group identification.

\paragraph{\bf Results} From Table~\ref{tab:group} we see that our learning process of the KPM gate is very stable and has very low standard deviation across training runs. Our simple method takes a few miliseconds to run and could be used for real-time group identification. On the Synthetic dataset, our approach significantly outperforms all state-of-the-art methods in terms of all metrics. On the Coffee Break dataset, our approach outperforms other methods in terms of Precision. and performs as well as the best (GCFF) in terms of F1 score. On the Cocktail Party dataset, our method performs slightly worse than the best (GCFF and GRUPO), but better than HVFF-ms.  Overall, our method outperforms all methods in terms of Precision and F1 score, and performs comparably to the best (GCFF) in terms of Recall. 

\paragraph{\bf Discussion} Our work improves over prior work in ways that are even more significant than our superior overall performance: \\{\emph {(1) Imitation leads to group detection:}} Most importantly, unlike prior work, our model is not trained explicitly for group detection. Instead, our focus is on behavioral imitation (learning a model that mimics human interaction). The surprising result is that, after training for behavior imitation, the output of KPM gate also performs successfully on the task of group detection. \\{\emph {(2) Data-driven approach, features, parameters:}} All methods we compare with rely on the assumption that people in groups position themselves in specific layouts (F-formations) developed in interactional socio-linguistics. By contrast, our models use no prior knowledge about layouts - we learn spatial layouts from data. Methods like GRUPO \cite{Vazquez2015} use hand-defined parameters, features (e.g, stride, lambda, gaussian mixtures), and expensive iterative mode-finding sub-algorithms. By contrast, our method has a single learned feature (KPM gate) and performs low-cost O($N^2$) unsupervised clustering. GRUPO uses the additional feature of lower body orientation, which our method does not.

In summary, we have demonstrated state-of-the-art performance using a simple, direct method without the use of hand-crafted parameters, features or layout assumptions. The remarkable \emph{emergence} of social perception in the form of group identification has been compared with various \emph{explicit} methods for this task.

\section{Conclusion}

In this work, we have presented \mgp{}, a deep neural network model of non-verbal and conversational interactions among multiple people and multiple groups. In the absence of real-world annotated data or simulators of multigroup face-to-face conversational behavior (\eg, speaking, listening, responding), we have designed our own social interaction simulator. While we consider discrete conversational actions among agents, we believe this is a necessary first step toward designing more complex models of real-world multigroup social intelligence. 

We have demonstrated the necessity and efficacy of each interpretable component of \mgp{}, by evaluating several ablative baselines on their ability to predict the next appropriate action. We have demonstrated \mgp{}'s superior performance in scenarios with static and dynamically evolving group assignments. We reiterate that training \mgp{} requires no explicit group annotations. Instead, its Kinesic-Proxemic-Message gate (KPM gate) learns to express social attention based on non-verbal cues, as a result of training for a socially intelligent policy. 

We have demonstrated the remarkable emergence of KPM gate's ability to identify groups in real-world data, and achieved state-of-the-art results with our direct, unsupervised method, without using complex layout assumptions or hand-defined parameters.

In a dearth of real-world data on multiagent multigroup conversational interactions, valuable future work would involve large-scale collection of non-verbal and verbal exchanges between multiple groups of people. The availability of more diverse data of human-to-human exchange (\eg, body orientation, gestures, facial expressions, \etc) would equip us to better design computational models like \mgp{}. Furthermore, such data would equip our models to better reason about social perception and propriety, by allowing the training of deep modules for a wider range of classes of behavioral actions, \eg, illustrators, emblems and attitudes \cite{vinciarelli2009social}. 

\vspace{3mm}

{\noindent\bf Acknowledgment:} This project was sponsored in part by JST CREST (JPMJCR14E1), NSF NRI (1637927) and IARPA (D17PC00340).
\pagebreak



\begin{thebibliography}{62}


\ifx \showCODEN    \undefined \def \showCODEN     #1{\unskip}     \fi
\ifx \showDOI      \undefined \def \showDOI       #1{#1}\fi
\ifx \showISBNx    \undefined \def \showISBNx     #1{\unskip}     \fi
\ifx \showISBNxiii \undefined \def \showISBNxiii  #1{\unskip}     \fi
\ifx \showISSN     \undefined \def \showISSN      #1{\unskip}     \fi
\ifx \showLCCN     \undefined \def \showLCCN      #1{\unskip}     \fi
\ifx \shownote     \undefined \def \shownote      #1{#1}          \fi
\ifx \showarticletitle \undefined \def \showarticletitle #1{#1}   \fi
\ifx \showURL      \undefined \def \showURL       {\relax}        \fi
\providecommand\bibfield[2]{#2}
\providecommand\bibinfo[2]{#2}
\providecommand\natexlab[1]{#1}
\providecommand\showeprint[2][]{arXiv:#2}

\bibitem[\protect\citeauthoryear{Alahi, Goel, Ramanathan, Robicquet, Fei-Fei,
  and Savarese}{Alahi et~al\mbox{.}}{2016}]%
        {Alahi2016}
\bibfield{author}{\bibinfo{person}{Alexandre Alahi}, \bibinfo{person}{Kratarth
  Goel}, \bibinfo{person}{Vignesh Ramanathan}, \bibinfo{person}{Alexandre
  Robicquet}, \bibinfo{person}{Li Fei-Fei}, {and} \bibinfo{person}{Silvio
  Savarese}.} \bibinfo{year}{2016}\natexlab{}.
\newblock \showarticletitle{Social LSTM: Human Trajectory Prediction in Crowded
  Spaces}. In \bibinfo{booktitle}{\emph{Proceeding of the IEEE Conference on
  Computer Vision and Pattern Recognition}}.
\newblock


\bibitem[\protect\citeauthoryear{Albrecht}{Albrecht}{2006}]%
        {albrecht2006social}
\bibfield{author}{\bibinfo{person}{Karl Albrecht}.}
  \bibinfo{year}{2006}\natexlab{}.
\newblock \bibinfo{booktitle}{\emph{Social intelligence: The new science of
  success}}.
\newblock \bibinfo{publisher}{John Wiley \& Sons}.
\newblock


\bibitem[\protect\citeauthoryear{Bazzani, Cristani, Tosato, Farenzena,
  Paggetti, Menegaz, and Murino}{Bazzani et~al\mbox{.}}{2013}]%
        {bazzani2013social}
\bibfield{author}{\bibinfo{person}{Loris Bazzani}, \bibinfo{person}{Marco
  Cristani}, \bibinfo{person}{Diego Tosato}, \bibinfo{person}{Michela
  Farenzena}, \bibinfo{person}{Giulia Paggetti}, \bibinfo{person}{Gloria
  Menegaz}, {and} \bibinfo{person}{Vittorio Murino}.}
  \bibinfo{year}{2013}\natexlab{}.
\newblock \showarticletitle{Social interactions by visual focus of attention in
  a three-dimensional environment}.
\newblock \bibinfo{journal}{\emph{Expert Systems}} \bibinfo{volume}{30},
  \bibinfo{number}{2} (\bibinfo{year}{2013}), \bibinfo{pages}{115--127}.
\newblock


\bibitem[\protect\citeauthoryear{Birdwhistell}{Birdwhistell}{1952}]%
        {birdwhistell1952introduction}
\bibfield{author}{\bibinfo{person}{Ray~L Birdwhistell}.}
  \bibinfo{year}{1952}\natexlab{}.
\newblock \bibinfo{booktitle}{\emph{Introduction to kinesics: An annotation
  system for analysis of body motion and gesture}}.
\newblock \bibinfo{publisher}{University of Louisville}.
\newblock


\bibitem[\protect\citeauthoryear{Brambilla, Ferrante, Birattari, and
  Dorigo}{Brambilla et~al\mbox{.}}{2013}]%
        {brambilla2013swarm}
\bibfield{author}{\bibinfo{person}{Manuele Brambilla}, \bibinfo{person}{Eliseo
  Ferrante}, \bibinfo{person}{Mauro Birattari}, {and} \bibinfo{person}{Marco
  Dorigo}.} \bibinfo{year}{2013}\natexlab{}.
\newblock \showarticletitle{Swarm robotics: a review from the swarm engineering
  perspective}.
\newblock \bibinfo{journal}{\emph{Swarm Intelligence}} \bibinfo{volume}{7},
  \bibinfo{number}{1} (\bibinfo{year}{2013}), \bibinfo{pages}{1--41}.
\newblock


\bibitem[\protect\citeauthoryear{Breazeal}{Breazeal}{2003}]%
        {breazeal2003toward}
\bibfield{author}{\bibinfo{person}{Cynthia Breazeal}.}
  \bibinfo{year}{2003}\natexlab{}.
\newblock \showarticletitle{Toward sociable robots}.
\newblock \bibinfo{journal}{\emph{Robotics and autonomous systems}}
  \bibinfo{volume}{42}, \bibinfo{number}{3-4} (\bibinfo{year}{2003}),
  \bibinfo{pages}{167--175}.
\newblock


\bibitem[\protect\citeauthoryear{Breazeal, Buchsbaum, Gray, Gatenby, and
  Blumberg}{Breazeal et~al\mbox{.}}{2005}]%
        {breazeal2005learning}
\bibfield{author}{\bibinfo{person}{Cynthia Breazeal}, \bibinfo{person}{Daphna
  Buchsbaum}, \bibinfo{person}{Jesse Gray}, \bibinfo{person}{David Gatenby},
  {and} \bibinfo{person}{Bruce Blumberg}.} \bibinfo{year}{2005}\natexlab{}.
\newblock \showarticletitle{Learning from and about others: Towards using
  imitation to bootstrap the social understanding of others by robots}.
\newblock \bibinfo{journal}{\emph{Artificial life}} \bibinfo{volume}{11},
  \bibinfo{number}{1-2} (\bibinfo{year}{2005}), \bibinfo{pages}{31--62}.
\newblock


\bibitem[\protect\citeauthoryear{Bronkhorst}{Bronkhorst}{2000}]%
        {bronkhorst2000cocktail}
\bibfield{author}{\bibinfo{person}{Adelbert~W Bronkhorst}.}
  \bibinfo{year}{2000}\natexlab{}.
\newblock \showarticletitle{The cocktail party phenomenon: A review of research
  on speech intelligibility in multiple-talker conditions}.
\newblock \bibinfo{journal}{\emph{Acta Acustica united with Acustica}}
  \bibinfo{volume}{86}, \bibinfo{number}{1} (\bibinfo{year}{2000}),
  \bibinfo{pages}{117--128}.
\newblock


\bibitem[\protect\citeauthoryear{Busoniu, Babu${\hat s}$ka, and
  Schutter}{Busoniu et~al\mbox{.}}{2008}]%
        {Busoniu2008}
\bibfield{author}{\bibinfo{person}{L. Busoniu}, \bibinfo{person}{R. Babu${\hat
  s}$ka}, {and} \bibinfo{person}{B.~D. Schutter}.}
  \bibinfo{year}{2008}\natexlab{}.
\newblock \showarticletitle{A Comprehensive Survey of Multiagent Reinforcement
  Learning}.
\newblock \bibinfo{journal}{\emph{IEEE Transactions on Systems, Man, and
  Cybernetics, Part C (Applications and Reviews)}} \bibinfo{volume}{38},
  \bibinfo{number}{2} (\bibinfo{date}{March} \bibinfo{year}{2008}),
  \bibinfo{pages}{156--172}.
\newblock
\showISSN{1094-6977}
\urldef\tempurl%
\url{https://doi.org/10.1109/TSMCC.2007.913919}
\showDOI{\tempurl}


\bibitem[\protect\citeauthoryear{Castellano, Leite, Pereira, Martinho, Paiva,
  and McOwan}{Castellano et~al\mbox{.}}{2012}]%
        {castellano2012detecting}
\bibfield{author}{\bibinfo{person}{Ginevra Castellano},
  \bibinfo{person}{Iolanda Leite}, \bibinfo{person}{Andre Pereira},
  \bibinfo{person}{Carlos Martinho}, \bibinfo{person}{Ana Paiva}, {and}
  \bibinfo{person}{Peter~W McOwan}.} \bibinfo{year}{2012}\natexlab{}.
\newblock \showarticletitle{Detecting engagement in HRI: An exploration of
  social and task-based context}. In \bibinfo{booktitle}{\emph{2012
  International Conference on Privacy, Security, Risk and Trust and 2012
  International Confernece on Social Computing}}. IEEE,
  \bibinfo{pages}{421--428}.
\newblock


\bibitem[\protect\citeauthoryear{Chung, G{\"{u}}l{\c{c}}ehre, Cho, and
  Bengio}{Chung et~al\mbox{.}}{2014}]%
        {Chung2014}
\bibfield{author}{\bibinfo{person}{Junyoung Chung},
  \bibinfo{person}{{\c{C}}aglar G{\"{u}}l{\c{c}}ehre},
  \bibinfo{person}{KyungHyun Cho}, {and} \bibinfo{person}{Yoshua Bengio}.}
  \bibinfo{year}{2014}\natexlab{}.
\newblock \showarticletitle{Empirical Evaluation of Gated Recurrent Neural
  Networks on Sequence Modeling}.
\newblock \bibinfo{journal}{\emph{CoRR}}  \bibinfo{volume}{abs/1412.3555}
  (\bibinfo{year}{2014}).
\newblock


\bibitem[\protect\citeauthoryear{Clevert, Unterthiner, and Hochreiter}{Clevert
  et~al\mbox{.}}{2015}]%
        {Clevert2015}
\bibfield{author}{\bibinfo{person}{Djork{-}Arn{\'{e}} Clevert},
  \bibinfo{person}{Thomas Unterthiner}, {and} \bibinfo{person}{Sepp
  Hochreiter}.} \bibinfo{year}{2015}\natexlab{}.
\newblock \showarticletitle{Fast and Accurate Deep Network Learning by
  Exponential Linear Units (ELUs)}.
\newblock \bibinfo{journal}{\emph{CoRR}}  \bibinfo{volume}{abs/1511.07289}
  (\bibinfo{year}{2015}).
\newblock


\bibitem[\protect\citeauthoryear{Colino, Buckingham, Cheng, van Donkelaar, and
  Binsted}{Colino et~al\mbox{.}}{2014}]%
        {colino2014tactile}
\bibfield{author}{\bibinfo{person}{Francisco~L Colino}, \bibinfo{person}{Gavin
  Buckingham}, \bibinfo{person}{Darian~T Cheng}, \bibinfo{person}{Paul van
  Donkelaar}, {and} \bibinfo{person}{Gordon Binsted}.}
  \bibinfo{year}{2014}\natexlab{}.
\newblock \showarticletitle{Tactile gating in a reaching and grasping task}.
\newblock \bibinfo{journal}{\emph{Physiological reports}} \bibinfo{volume}{2},
  \bibinfo{number}{3} (\bibinfo{year}{2014}).
\newblock


\bibitem[\protect\citeauthoryear{Courbariaux and Bengio}{Courbariaux and
  Bengio}{2016}]%
        {Courbariaux2016}
\bibfield{author}{\bibinfo{person}{Matthieu Courbariaux} {and}
  \bibinfo{person}{Yoshua Bengio}.} \bibinfo{year}{2016}\natexlab{}.
\newblock \showarticletitle{BinaryNet: Training Deep Neural Networks with
  Weights and Activations Constrained to +1 or -1}.
\newblock \bibinfo{journal}{\emph{CoRR}}  \bibinfo{volume}{abs/1602.02830}
  (\bibinfo{year}{2016}).
\newblock


\bibitem[\protect\citeauthoryear{Cristani, Bazzani, Paggetti, Fossati, Tosato,
  Bue, Menegaz, and Murino}{Cristani et~al\mbox{.}}{2011}]%
        {Cristani2011}
\bibfield{author}{\bibinfo{person}{Marco Cristani}, \bibinfo{person}{Loris
  Bazzani}, \bibinfo{person}{Giulia Paggetti}, \bibinfo{person}{Andrea
  Fossati}, \bibinfo{person}{Diego Tosato}, \bibinfo{person}{Alessio~Del Bue},
  \bibinfo{person}{Gloria Menegaz}, {and} \bibinfo{person}{Vittorio Murino}.}
  \bibinfo{year}{2011}\natexlab{}.
\newblock \showarticletitle{Social interaction discovery by statistical
  analysis of F-formations}. In \bibinfo{booktitle}{\emph{British Machine
  Vision Conference}}. \bibinfo{pages}{23.1--23.12}.
\newblock


\bibitem[\protect\citeauthoryear{Cristani, Murino, and Vinciarelli}{Cristani
  et~al\mbox{.}}{2010}]%
        {cristani2010socially}
\bibfield{author}{\bibinfo{person}{Marco Cristani}, \bibinfo{person}{Vittorio
  Murino}, {and} \bibinfo{person}{Alessandro Vinciarelli}.}
  \bibinfo{year}{2010}\natexlab{}.
\newblock \showarticletitle{Socially intelligent surveillance and monitoring:
  Analysing social dimensions of physical space}. In
  \bibinfo{booktitle}{\emph{2010 IEEE Computer Society Conference on Computer
  Vision and Pattern Recognition-Workshops}}. IEEE, \bibinfo{pages}{51--58}.
\newblock


\bibitem[\protect\citeauthoryear{Cristani, Raghavendra, Del~Bue, and
  Murino}{Cristani et~al\mbox{.}}{2013}]%
        {cristani2013human}
\bibfield{author}{\bibinfo{person}{Marco Cristani}, \bibinfo{person}{Ramya
  Raghavendra}, \bibinfo{person}{Alessio Del~Bue}, {and}
  \bibinfo{person}{Vittorio Murino}.} \bibinfo{year}{2013}\natexlab{}.
\newblock \showarticletitle{Human behavior analysis in video surveillance: A
  social signal processing perspective}.
\newblock \bibinfo{journal}{\emph{Neurocomputing}}  \bibinfo{volume}{100}
  (\bibinfo{year}{2013}), \bibinfo{pages}{86--97}.
\newblock


\bibitem[\protect\citeauthoryear{Cromwell, Mears, Wan, and Boutros}{Cromwell
  et~al\mbox{.}}{2008}]%
        {cromwell2008sensory}
\bibfield{author}{\bibinfo{person}{Howard~C Cromwell}, \bibinfo{person}{Ryan~P
  Mears}, \bibinfo{person}{Li Wan}, {and} \bibinfo{person}{Nash~N Boutros}.}
  \bibinfo{year}{2008}\natexlab{}.
\newblock \showarticletitle{Sensory gating: a translational effort from basic
  to clinical science}.
\newblock \bibinfo{journal}{\emph{Clinical EEG and Neuroscience}}
  \bibinfo{volume}{39}, \bibinfo{number}{2} (\bibinfo{year}{2008}),
  \bibinfo{pages}{69--72}.
\newblock


\bibitem[\protect\citeauthoryear{Dautenhahn}{Dautenhahn}{2007}]%
        {dautenhahn2007socially}
\bibfield{author}{\bibinfo{person}{Kerstin Dautenhahn}.}
  \bibinfo{year}{2007}\natexlab{}.
\newblock \showarticletitle{Socially intelligent robots: dimensions of
  human--robot interaction}.
\newblock \bibinfo{journal}{\emph{Philosophical transactions of the royal
  society B: Biological sciences}} \bibinfo{volume}{362},
  \bibinfo{number}{1480} (\bibinfo{year}{2007}), \bibinfo{pages}{679--704}.
\newblock


\bibitem[\protect\citeauthoryear{Davies and Stone}{Davies and Stone}{1995}]%
        {davies1995mental}
\bibfield{author}{\bibinfo{person}{Martin Davies} {and} \bibinfo{person}{Tony
  Stone}.} \bibinfo{year}{1995}\natexlab{}.
\newblock \showarticletitle{Mental Simulation: Evaluations and
  Applications-Reading in Mind and Language}.
\newblock  (\bibinfo{year}{1995}).
\newblock


\bibitem[\protect\citeauthoryear{Edward}{Edward}{1966}]%
        {edward1966hall}
\bibfield{author}{\bibinfo{person}{T Edward}.} \bibinfo{year}{1966}\natexlab{}.
\newblock \bibinfo{title}{Hall, The Hidden Dimension}.
\newblock
\newblock


\bibitem[\protect\citeauthoryear{Ester, Kriegel, Sander, and Xu}{Ester
  et~al\mbox{.}}{1996}]%
        {Ester1996}
\bibfield{author}{\bibinfo{person}{Martin Ester}, \bibinfo{person}{Hans-Peter
  Kriegel}, \bibinfo{person}{J\"{o}rg Sander}, {and} \bibinfo{person}{Xiaowei
  Xu}.} \bibinfo{year}{1996}\natexlab{}.
\newblock \showarticletitle{A Density-based Algorithm for Discovering Clusters
  a Density-based Algorithm for Discovering Clusters in Large Spatial Databases
  with Noise}. In \bibinfo{booktitle}{\emph{Proceedings of the International
  Conference on Knowledge Discovery and Data Mining}}.
  \bibinfo{pages}{226--231}.
\newblock


\bibitem[\protect\citeauthoryear{Fiore, Wiltshire, Lobato, Jentsch, Huang, and
  Axelrod}{Fiore et~al\mbox{.}}{2013}]%
        {fiore2013toward}
\bibfield{author}{\bibinfo{person}{Stephen~M Fiore}, \bibinfo{person}{Travis~J
  Wiltshire}, \bibinfo{person}{Emilio~JC Lobato}, \bibinfo{person}{Florian~G
  Jentsch}, \bibinfo{person}{Wesley~H Huang}, {and} \bibinfo{person}{Benjamin
  Axelrod}.} \bibinfo{year}{2013}\natexlab{}.
\newblock \showarticletitle{Toward understanding social cues and signals in
  human--robot interaction: effects of robot gaze and proxemic behavior}.
\newblock \bibinfo{journal}{\emph{Frontiers in psychology}}
  \bibinfo{volume}{4} (\bibinfo{year}{2013}), \bibinfo{pages}{859}.
\newblock


\bibitem[\protect\citeauthoryear{Foerster, Assael, de~Freitas, and
  Whiteson}{Foerster et~al\mbox{.}}{2016}]%
        {Foerster2016}
\bibfield{author}{\bibinfo{person}{Jakob Foerster}, \bibinfo{person}{Yannis~M.
  Assael}, \bibinfo{person}{Nando de Freitas}, {and} \bibinfo{person}{Shimon
  Whiteson}.} \bibinfo{year}{2016}\natexlab{}.
\newblock \showarticletitle{Learning to Communicate with Deep Multi-Agent
  Reinforcement Learning}. In \bibinfo{booktitle}{\emph{Proceeding of the
  Advances in Neural Information Processing Systems}}.
  \bibinfo{pages}{2137--2145}.
\newblock


\bibitem[\protect\citeauthoryear{Foerster, Farquhar, Afouras, Nardelli, and
  Whiteson}{Foerster et~al\mbox{.}}{2017}]%
        {Foerster2017}
\bibfield{author}{\bibinfo{person}{Jakob~N. Foerster}, \bibinfo{person}{Gregory
  Farquhar}, \bibinfo{person}{Triantafyllos Afouras}, \bibinfo{person}{Nantas
  Nardelli}, {and} \bibinfo{person}{Shimon Whiteson}.}
  \bibinfo{year}{2017}\natexlab{}.
\newblock \showarticletitle{Counterfactual Multi-Agent Policy Gradients}.
\newblock \bibinfo{journal}{\emph{CoRR}}  \bibinfo{volume}{abs/1705.08926}
  (\bibinfo{year}{2017}).
\newblock
\urldef\tempurl%
\url{http://arxiv.org/abs/1705.08926}
\showURL{%
\tempurl}


\bibitem[\protect\citeauthoryear{Freedman, Adler, Gerhardt, Waldo, Baker, Rose,
  Drebing, Nagamoto, Bickford-Wimer, and Franks}{Freedman
  et~al\mbox{.}}{1987}]%
        {freedman1987neurobiological}
\bibfield{author}{\bibinfo{person}{Robert Freedman},
  \bibinfo{person}{Lawrence~E Adler}, \bibinfo{person}{Greg~A Gerhardt},
  \bibinfo{person}{Merilyne Waldo}, \bibinfo{person}{Neil Baker},
  \bibinfo{person}{Greg~M Rose}, \bibinfo{person}{Carla Drebing},
  \bibinfo{person}{Herbert Nagamoto}, \bibinfo{person}{Paula Bickford-Wimer},
  {and} \bibinfo{person}{Ronald Franks}.} \bibinfo{year}{1987}\natexlab{}.
\newblock \showarticletitle{Neurobiological studies of sensory gating in
  schizophrenia}.
\newblock \bibinfo{journal}{\emph{Schizophrenia bulletin}}
  \bibinfo{volume}{13}, \bibinfo{number}{4} (\bibinfo{year}{1987}),
  \bibinfo{pages}{669--678}.
\newblock


\bibitem[\protect\citeauthoryear{Gatica-Perez}{Gatica-Perez}{2006}]%
        {Gatica2006}
\bibfield{author}{\bibinfo{person}{D. Gatica-Perez}.}
  \bibinfo{year}{2006}\natexlab{}.
\newblock \showarticletitle{Analyzing Group Interactions in Conversations: a
  Review}. In \bibinfo{booktitle}{\emph{Proceedings of the International
  Conference on Multisensor Fusion and Integration for Intelligent Systems}}.
  \bibinfo{pages}{41--46}.
\newblock


\bibitem[\protect\citeauthoryear{Gatica-Perez, McCowan, Zhang, and
  Bengio}{Gatica-Perez et~al\mbox{.}}{2005}]%
        {gatica2005detecting}
\bibfield{author}{\bibinfo{person}{Daniel Gatica-Perez}, \bibinfo{person}{L
  McCowan}, \bibinfo{person}{Dong Zhang}, {and} \bibinfo{person}{Samy Bengio}.}
  \bibinfo{year}{2005}\natexlab{}.
\newblock \showarticletitle{Detecting group interest-level in meetings}. In
  \bibinfo{booktitle}{\emph{Proceedings.(ICASSP'05). IEEE International
  Conference on Acoustics, Speech, and Signal Processing, 2005.}},
  Vol.~\bibinfo{volume}{1}. IEEE, \bibinfo{pages}{I--489}.
\newblock


\bibitem[\protect\citeauthoryear{Goleman}{Goleman}{2007}]%
        {goleman2007social}
\bibfield{author}{\bibinfo{person}{Daniel Goleman}.}
  \bibinfo{year}{2007}\natexlab{}.
\newblock \bibinfo{booktitle}{\emph{Social intelligence}}.
\newblock \bibinfo{publisher}{Random house}.
\newblock


\bibitem[\protect\citeauthoryear{Heylen, Reidsma, and Ordelman}{Heylen
  et~al\mbox{.}}{2006}]%
        {heylen2006annotating}
\bibfield{author}{\bibinfo{person}{Dirk Heylen}, \bibinfo{person}{Dennis
  Reidsma}, {and} \bibinfo{person}{Roeland Ordelman}.}
  \bibinfo{year}{2006}\natexlab{}.
\newblock \showarticletitle{Annotating state of mind in meeting data}. In
  \bibinfo{booktitle}{\emph{The Workshop Programme Corpora for Research on
  Emotion and Affect}}. \bibinfo{pages}{84}.
\newblock


\bibitem[\protect\citeauthoryear{Hillyard and Mangun}{Hillyard and
  Mangun}{1987}]%
        {hillyard1987sensory}
\bibfield{author}{\bibinfo{person}{SA Hillyard} {and} \bibinfo{person}{GR
  Mangun}.} \bibinfo{year}{1987}\natexlab{}.
\newblock \showarticletitle{Sensory gating as a physiological mechanism for
  visual selective attention.}
\newblock \bibinfo{journal}{\emph{Electroencephalography and clinical
  neurophysiology. Supplement}}  \bibinfo{volume}{40} (\bibinfo{year}{1987}),
  \bibinfo{pages}{61--67}.
\newblock


\bibitem[\protect\citeauthoryear{Hung and Kr\"{o}se}{Hung and
  Kr\"{o}se}{2011}]%
        {Hung2011}
\bibfield{author}{\bibinfo{person}{Hayley Hung} {and} \bibinfo{person}{Ben
  Kr\"{o}se}.} \bibinfo{year}{2011}\natexlab{}.
\newblock \showarticletitle{Detecting F-formations As Dominant Sets}. In
  \bibinfo{booktitle}{\emph{Proceedings of the International Conference on
  Multimodal Interfaces}}. \bibinfo{pages}{231--238}.
\newblock


\bibitem[\protect\citeauthoryear{Katzenmaier, Stiefelhagen, and
  Schultz}{Katzenmaier et~al\mbox{.}}{2004}]%
        {katzenmaier2004identifying}
\bibfield{author}{\bibinfo{person}{Michael Katzenmaier},
  \bibinfo{person}{Rainer Stiefelhagen}, {and} \bibinfo{person}{Tanja
  Schultz}.} \bibinfo{year}{2004}\natexlab{}.
\newblock \showarticletitle{Identifying the addressee in human-human-robot
  interactions based on head pose and speech}. In
  \bibinfo{booktitle}{\emph{Proceedings of the 6th international conference on
  Multimodal interfaces}}. ACM, \bibinfo{pages}{144--151}.
\newblock


\bibitem[\protect\citeauthoryear{Kendon}{Kendon}{1990}]%
        {Kendon1990}
\bibfield{author}{\bibinfo{person}{A. Kendon}.}
  \bibinfo{year}{1990}\natexlab{}.
\newblock \bibinfo{booktitle}{\emph{Conducting interaction: {P}attern of
  behavior in focused encounter}}.
\newblock \bibinfo{publisher}{Cambridge University Press}.
\newblock


\bibitem[\protect\citeauthoryear{Kingma and Ba}{Kingma and Ba}{2014}]%
        {Diederik2014}
\bibfield{author}{\bibinfo{person}{Diederik~P. Kingma} {and}
  \bibinfo{person}{Jimmy Ba}.} \bibinfo{year}{2014}\natexlab{}.
\newblock \showarticletitle{Adam: {A} Method for Stochastic Optimization}.
\newblock \bibinfo{journal}{\emph{CoRR}}  \bibinfo{volume}{abs/1412.6980}
  (\bibinfo{year}{2014}).
\newblock
\urldef\tempurl%
\url{http://arxiv.org/abs/1412.6980}
\showURL{%
\tempurl}


\bibitem[\protect\citeauthoryear{Le, Yue, and Carr}{Le et~al\mbox{.}}{2017}]%
        {Le2017}
\bibfield{author}{\bibinfo{person}{Hoang~M. Le}, \bibinfo{person}{Yisong Yue},
  {and} \bibinfo{person}{Peter Carr}.} \bibinfo{year}{2017}\natexlab{}.
\newblock \showarticletitle{Coordinated Multi-Agent Imitation Learning}. In
  \bibinfo{booktitle}{\emph{Proceeding of the International Conference on
  Machine Learning}}.
\newblock


\bibitem[\protect\citeauthoryear{Lee, Choi, Vernaza, Choy, Torr, and
  Chandraker}{Lee et~al\mbox{.}}{2017}]%
        {Lee2017}
\bibfield{author}{\bibinfo{person}{Namhoon Lee}, \bibinfo{person}{Wongun Choi},
  \bibinfo{person}{Paul Vernaza}, \bibinfo{person}{Christopher~B. Choy},
  \bibinfo{person}{Philip H.~S. Torr}, {and} \bibinfo{person}{Manmohan
  Chandraker}.} \bibinfo{year}{2017}\natexlab{}.
\newblock \showarticletitle{DESIRE: Distant Future Prediction in Dynamic Scenes
  With Interacting Agents}. In \bibinfo{booktitle}{\emph{Proceedings of the
  IEEE Conference on Computer Vision and Pattern Recognition}}.
\newblock


\bibitem[\protect\citeauthoryear{Lowe, Wu, Tamar, Harb, Abbeel, and
  Mordatch}{Lowe et~al\mbox{.}}{2017}]%
        {Lowe2017}
\bibfield{author}{\bibinfo{person}{Ryan Lowe}, \bibinfo{person}{Yi Wu},
  \bibinfo{person}{Aviv Tamar}, \bibinfo{person}{Jean Harb},
  \bibinfo{person}{Pieter Abbeel}, {and} \bibinfo{person}{Igor Mordatch}.}
  \bibinfo{year}{2017}\natexlab{}.
\newblock \showarticletitle{Multi-Agent Actor-Critic for Mixed
  Cooperative-Competitive Environments}.
\newblock \bibinfo{journal}{\emph{CoRR}}  \bibinfo{volume}{abs/1706.02275}
  (\bibinfo{year}{2017}).
\newblock
\urldef\tempurl%
\url{http://arxiv.org/abs/1706.02275}
\showURL{%
\tempurl}


\bibitem[\protect\citeauthoryear{Marsella, Pynadath, and Read}{Marsella
  et~al\mbox{.}}{2004}]%
        {marsella2004psychsim}
\bibfield{author}{\bibinfo{person}{Stacy~C Marsella}, \bibinfo{person}{David~V
  Pynadath}, {and} \bibinfo{person}{Stephen~J Read}.}
  \bibinfo{year}{2004}\natexlab{}.
\newblock \showarticletitle{PsychSim: Agent-based modeling of social
  interactions and influence}. In \bibinfo{booktitle}{\emph{Proceedings of the
  international conference on cognitive modeling}}, Vol.~\bibinfo{volume}{36}.
  \bibinfo{pages}{243--248}.
\newblock


\bibitem[\protect\citeauthoryear{Mast}{Mast}{2002}]%
        {mast2002dominance}
\bibfield{author}{\bibinfo{person}{Marianne~Schmid Mast}.}
  \bibinfo{year}{2002}\natexlab{}.
\newblock \showarticletitle{Dominance as expressed and inferred through
  speaking time: A meta-analysis}.
\newblock \bibinfo{journal}{\emph{Human Communication Research}}
  \bibinfo{volume}{28}, \bibinfo{number}{3} (\bibinfo{year}{2002}),
  \bibinfo{pages}{420--450}.
\newblock


\bibitem[\protect\citeauthoryear{Mehrabian and Ferris}{Mehrabian and
  Ferris}{1967}]%
        {mehrabian1967inference}
\bibfield{author}{\bibinfo{person}{Albert Mehrabian} {and}
  \bibinfo{person}{Susan~R Ferris}.} \bibinfo{year}{1967}\natexlab{}.
\newblock \showarticletitle{Inference of attitudes from nonverbal communication
  in two channels.}
\newblock \bibinfo{journal}{\emph{Journal of consulting psychology}}
  \bibinfo{volume}{31}, \bibinfo{number}{3} (\bibinfo{year}{1967}),
  \bibinfo{pages}{248}.
\newblock


\bibitem[\protect\citeauthoryear{Meltzoff and Decety}{Meltzoff and
  Decety}{2003}]%
        {meltzoff2003imitation}
\bibfield{author}{\bibinfo{person}{Andrew~N Meltzoff} {and}
  \bibinfo{person}{Jean Decety}.} \bibinfo{year}{2003}\natexlab{}.
\newblock \showarticletitle{What imitation tells us about social cognition: a
  rapprochement between developmental psychology and cognitive neuroscience}.
\newblock \bibinfo{journal}{\emph{Philosophical Transactions of the Royal
  Society of London. Series B: Biological Sciences}} \bibinfo{volume}{358},
  \bibinfo{number}{1431} (\bibinfo{year}{2003}), \bibinfo{pages}{491--500}.
\newblock


\bibitem[\protect\citeauthoryear{Mordatch and Abbeel}{Mordatch and
  Abbeel}{2017}]%
        {Mordatch2017}
\bibfield{author}{\bibinfo{person}{Igor Mordatch} {and} \bibinfo{person}{Pieter
  Abbeel}.} \bibinfo{year}{2017}\natexlab{}.
\newblock \showarticletitle{Emergence of Grounded Compositional Language in
  Multi-Agent Populations}.
\newblock \bibinfo{journal}{\emph{CoRR}}  \bibinfo{volume}{abs/1703.04908}
  (\bibinfo{year}{2017}).
\newblock
\urldef\tempurl%
\url{http://arxiv.org/abs/1703.04908}
\showURL{%
\tempurl}


\bibitem[\protect\citeauthoryear{Peng, Yuan, Wen, Yang, Tang, Long, and
  Wang}{Peng et~al\mbox{.}}{2017}]%
        {Peng2017}
\bibfield{author}{\bibinfo{person}{Peng Peng}, \bibinfo{person}{Quan Yuan},
  \bibinfo{person}{Ying Wen}, \bibinfo{person}{Yaodong Yang},
  \bibinfo{person}{Zhenkun Tang}, \bibinfo{person}{Haitao Long}, {and}
  \bibinfo{person}{Jun Wang}.} \bibinfo{year}{2017}\natexlab{}.
\newblock \showarticletitle{Multiagent Bidirectionally-Coordinated Nets for
  Learning to Play StarCraft Combat Games}.
\newblock \bibinfo{journal}{\emph{CoRR}}  \bibinfo{volume}{abs/1703.10069}
  (\bibinfo{year}{2017}).
\newblock
\urldef\tempurl%
\url{http://arxiv.org/abs/1703.10069}
\showURL{%
\tempurl}


\bibitem[\protect\citeauthoryear{Robicquet, Sadeghian, Alahi, and
  Savarese}{Robicquet et~al\mbox{.}}{2016}]%
        {Robicquet2016}
\bibfield{author}{\bibinfo{person}{Alexandre Robicquet}, \bibinfo{person}{Amir
  Sadeghian}, \bibinfo{person}{Alexandre Alahi}, {and} \bibinfo{person}{Silvio
  Savarese}.} \bibinfo{year}{2016}\natexlab{}.
\newblock \showarticletitle{Learning Social Etiquette: Human Trajectory
  Understanding In Crowded Scenes}. In \bibinfo{booktitle}{\emph{Proceeding of
  the European Conference on Computer Vision}}. \bibinfo{pages}{549--565}.
\newblock


\bibitem[\protect\citeauthoryear{Salam, Celiktutan, Hupont, Gunes, and
  Chetouani}{Salam et~al\mbox{.}}{2016}]%
        {salam2016fully}
\bibfield{author}{\bibinfo{person}{Hanan Salam}, \bibinfo{person}{Oya
  Celiktutan}, \bibinfo{person}{Isabelle Hupont}, \bibinfo{person}{Hatice
  Gunes}, {and} \bibinfo{person}{Mohamed Chetouani}.}
  \bibinfo{year}{2016}\natexlab{}.
\newblock \showarticletitle{Fully automatic analysis of engagement and its
  relationship to personality in human-robot interactions}.
\newblock \bibinfo{journal}{\emph{IEEE Access}}  \bibinfo{volume}{5}
  (\bibinfo{year}{2016}), \bibinfo{pages}{705--721}.
\newblock


\bibitem[\protect\citeauthoryear{Sanghvi, Castellano, Leite, Pereira, McOwan,
  and Paiva}{Sanghvi et~al\mbox{.}}{2011}]%
        {sanghvi2011automatic}
\bibfield{author}{\bibinfo{person}{Jyotirmay Sanghvi}, \bibinfo{person}{Ginevra
  Castellano}, \bibinfo{person}{Iolanda Leite}, \bibinfo{person}{Andr{\'e}
  Pereira}, \bibinfo{person}{Peter~W McOwan}, {and} \bibinfo{person}{Ana
  Paiva}.} \bibinfo{year}{2011}\natexlab{}.
\newblock \showarticletitle{Automatic analysis of affective postures and body
  motion to detect engagement with a game companion}. In
  \bibinfo{booktitle}{\emph{Proceedings of the 6th international conference on
  Human-robot interaction}}. ACM, \bibinfo{pages}{305--312}.
\newblock


\bibitem[\protect\citeauthoryear{Sanghvi, Nagavalli, and Sycara}{Sanghvi
  et~al\mbox{.}}{2017}]%
        {sanghvi2017exploiting}
\bibfield{author}{\bibinfo{person}{Navyata Sanghvi}, \bibinfo{person}{Sasanka
  Nagavalli}, {and} \bibinfo{person}{Katia Sycara}.}
  \bibinfo{year}{2017}\natexlab{}.
\newblock \showarticletitle{Exploiting robotic swarm characteristics for
  adversarial subversion in coverage tasks}. In
  \bibinfo{booktitle}{\emph{Proceedings of the 16th Conference on Autonomous
  Agents and MultiAgent Systems}}. International Foundation for Autonomous
  Agents and Multiagent Systems, \bibinfo{pages}{511--519}.
\newblock


\bibitem[\protect\citeauthoryear{Serban, Sordoni, Bengio, Courville, and
  Pineau}{Serban et~al\mbox{.}}{2016}]%
        {serban2016building}
\bibfield{author}{\bibinfo{person}{Iulian~V Serban},
  \bibinfo{person}{Alessandro Sordoni}, \bibinfo{person}{Yoshua Bengio},
  \bibinfo{person}{Aaron Courville}, {and} \bibinfo{person}{Joelle Pineau}.}
  \bibinfo{year}{2016}\natexlab{}.
\newblock \showarticletitle{Building end-to-end dialogue systems using
  generative hierarchical neural network models}. In
  \bibinfo{booktitle}{\emph{Thirtieth AAAI Conference on Artificial
  Intelligence}}.
\newblock


\bibitem[\protect\citeauthoryear{Setti, Lanz, Ferrario, Murino, and
  Cristani}{Setti et~al\mbox{.}}{2013}]%
        {Setti2013}
\bibfield{author}{\bibinfo{person}{F. Setti}, \bibinfo{person}{O. Lanz},
  \bibinfo{person}{R. Ferrario}, \bibinfo{person}{V. Murino}, {and}
  \bibinfo{person}{M. Cristani}.} \bibinfo{year}{2013}\natexlab{}.
\newblock \showarticletitle{Multi-scale f-formation discovery for group
  detection}. In \bibinfo{booktitle}{\emph{Proceedings of the International
  Conference on Image Processing}}. \bibinfo{pages}{3547--3551}.
\newblock


\bibitem[\protect\citeauthoryear{Setti, Russell, Bassetti, and Cristani}{Setti
  et~al\mbox{.}}{2015}]%
        {Setti2015}
\bibfield{author}{\bibinfo{person}{Francesco Setti}, \bibinfo{person}{Chris
  Russell}, \bibinfo{person}{Chiara Bassetti}, {and} \bibinfo{person}{Marco
  Cristani}.} \bibinfo{year}{2015}\natexlab{}.
\newblock \showarticletitle{F-Formation Detection: Individuating Free-Standing
  Conversational Groups in Images}.
\newblock \bibinfo{journal}{\emph{PLOS ONE}} \bibinfo{volume}{10},
  \bibinfo{number}{5} (\bibinfo{year}{2015}), \bibinfo{pages}{1--26}.
\newblock


\bibitem[\protect\citeauthoryear{Shapiro, Caldwell, and Sorensen}{Shapiro
  et~al\mbox{.}}{1997}]%
        {shapiro1997personal}
\bibfield{author}{\bibinfo{person}{Kimron~L Shapiro}, \bibinfo{person}{Judy
  Caldwell}, {and} \bibinfo{person}{Robyn~E Sorensen}.}
  \bibinfo{year}{1997}\natexlab{}.
\newblock \showarticletitle{Personal names and the attentional blink: A visual"
  cocktail party" effect.}
\newblock \bibinfo{journal}{\emph{Journal of Experimental Psychology: Human
  Perception and Performance}} \bibinfo{volume}{23}, \bibinfo{number}{2}
  (\bibinfo{year}{1997}), \bibinfo{pages}{504}.
\newblock


\bibitem[\protect\citeauthoryear{Singh, Kearns, Litman, and Walker}{Singh
  et~al\mbox{.}}{2000}]%
        {singh2000reinforcement}
\bibfield{author}{\bibinfo{person}{Satinder~P Singh},
  \bibinfo{person}{Michael~J Kearns}, \bibinfo{person}{Diane~J Litman}, {and}
  \bibinfo{person}{Marilyn~A Walker}.} \bibinfo{year}{2000}\natexlab{}.
\newblock \showarticletitle{Reinforcement learning for spoken dialogue
  systems}. In \bibinfo{booktitle}{\emph{Advances in Neural Information
  Processing Systems}}. \bibinfo{pages}{956--962}.
\newblock


\bibitem[\protect\citeauthoryear{Stone and Veloso}{Stone and Veloso}{2000}]%
        {Stone2000}
\bibfield{author}{\bibinfo{person}{Peter Stone} {and} \bibinfo{person}{Manuela
  Veloso}.} \bibinfo{year}{2000}\natexlab{}.
\newblock \showarticletitle{Multiagent Systems: A Survey from a Machine
  Learning Perspective}.
\newblock \bibinfo{journal}{\emph{Autonomous Robots}} \bibinfo{volume}{8},
  \bibinfo{number}{3} (\bibinfo{date}{01 Jun} \bibinfo{year}{2000}),
  \bibinfo{pages}{345--383}.
\newblock
\showISSN{1573-7527}
\urldef\tempurl%
\url{https://doi.org/10.1023/A:1008942012299}
\showDOI{\tempurl}


\bibitem[\protect\citeauthoryear{Sukhbaatar, Szlam, and Fergus}{Sukhbaatar
  et~al\mbox{.}}{2016}]%
        {Sukhbaatar2016}
\bibfield{author}{\bibinfo{person}{Sainbayar Sukhbaatar},
  \bibinfo{person}{Arthur Szlam}, {and} \bibinfo{person}{Rob Fergus}.}
  \bibinfo{year}{2016}\natexlab{}.
\newblock \showarticletitle{Learning Multiagent Communication with
  Backpropagation}. In \bibinfo{booktitle}{\emph{Proceeding of the Advances in
  Neural Information Processing Systems}}. \bibinfo{pages}{2244--2252}.
\newblock


\bibitem[\protect\citeauthoryear{Swofford, Peruzzi, V{\'a}zquez,
  Mart{\'\i}n-Mart{\'\i}n, and Savarese}{Swofford et~al\mbox{.}}{2019}]%
        {swofford2019dante}
\bibfield{author}{\bibinfo{person}{Mason Swofford},
  \bibinfo{person}{John~Charles Peruzzi}, \bibinfo{person}{Marynel
  V{\'a}zquez}, \bibinfo{person}{Roberto Mart{\'\i}n-Mart{\'\i}n}, {and}
  \bibinfo{person}{Silvio Savarese}.} \bibinfo{year}{2019}\natexlab{}.
\newblock \showarticletitle{DANTE: Deep Affinity Network for Clustering
  Conversational Interactants}.
\newblock \bibinfo{journal}{\emph{arXiv preprint arXiv:1907.12910}}
  (\bibinfo{year}{2019}).
\newblock


\bibitem[\protect\citeauthoryear{Vascon, Mequanint, Cristani, Hung, Pelillo,
  and Murino}{Vascon et~al\mbox{.}}{2014}]%
        {vascon2014game}
\bibfield{author}{\bibinfo{person}{Sebastiano Vascon},
  \bibinfo{person}{Eyasu~Zemene Mequanint}, \bibinfo{person}{Marco Cristani},
  \bibinfo{person}{Hayley Hung}, \bibinfo{person}{Marcello Pelillo}, {and}
  \bibinfo{person}{Vittorio Murino}.} \bibinfo{year}{2014}\natexlab{}.
\newblock \showarticletitle{A game-theoretic probabilistic approach for
  detecting conversational groups}. In \bibinfo{booktitle}{\emph{Asian
  conference on computer vision}}. Springer, \bibinfo{pages}{658--675}.
\newblock


\bibitem[\protect\citeauthoryear{Vazquez, Steinfeld, and Hudson}{Vazquez
  et~al\mbox{.}}{2015}]%
        {Vazquez2015}
\bibfield{author}{\bibinfo{person}{Marynel Vazquez}, \bibinfo{person}{Aaron
  Steinfeld}, {and} \bibinfo{person}{Scott~E. Hudson}.}
  \bibinfo{year}{2015}\natexlab{}.
\newblock \showarticletitle{Parallel Detection of Conversational Groups of
  Free-Standing People and Tracking of their Lower-Body Orientation}. In
  \bibinfo{booktitle}{\emph{Proceedings of the IEEE/RSJ International
  Conference on Intelligent Robots and Systems}}.
\newblock


\bibitem[\protect\citeauthoryear{Vinciarelli, Pantic, and Bourlard}{Vinciarelli
  et~al\mbox{.}}{2009}]%
        {vinciarelli2009social}
\bibfield{author}{\bibinfo{person}{Alessandro Vinciarelli},
  \bibinfo{person}{Maja Pantic}, {and} \bibinfo{person}{Herv{\'e} Bourlard}.}
  \bibinfo{year}{2009}\natexlab{}.
\newblock \showarticletitle{Social signal processing: Survey of an emerging
  domain}.
\newblock \bibinfo{journal}{\emph{Image and vision computing}}
  \bibinfo{volume}{27}, \bibinfo{number}{12} (\bibinfo{year}{2009}),
  \bibinfo{pages}{1743--1759}.
\newblock


\bibitem[\protect\citeauthoryear{Wagner and Arkin}{Wagner and Arkin}{2011}]%
        {wagner2011acting}
\bibfield{author}{\bibinfo{person}{Alan~R Wagner} {and}
  \bibinfo{person}{Ronald~C Arkin}.} \bibinfo{year}{2011}\natexlab{}.
\newblock \showarticletitle{Acting deceptively: Providing robots with the
  capacity for deception}.
\newblock \bibinfo{journal}{\emph{International Journal of Social Robotics}}
  \bibinfo{volume}{3}, \bibinfo{number}{1} (\bibinfo{year}{2011}),
  \bibinfo{pages}{5--26}.
\newblock


\bibitem[\protect\citeauthoryear{Yun, Lee, Park, Kim, and Kim}{Yun
  et~al\mbox{.}}{2018}]%
        {yun2018automatic}
\bibfield{author}{\bibinfo{person}{Woo-Han Yun}, \bibinfo{person}{Dongjin Lee},
  \bibinfo{person}{Chankyu Park}, \bibinfo{person}{Jaehong Kim}, {and}
  \bibinfo{person}{Junmo Kim}.} \bibinfo{year}{2018}\natexlab{}.
\newblock \showarticletitle{Automatic Recognition of Children Engagement from
  Facial Video using Convolutional Neural Networks}.
\newblock \bibinfo{journal}{\emph{IEEE Transactions on Affective Computing}}
  (\bibinfo{year}{2018}).
\newblock


\bibitem[\protect\citeauthoryear{Zen, Lepri, Ricci, and Lanz}{Zen
  et~al\mbox{.}}{2010}]%
        {Zen2010}
\bibfield{author}{\bibinfo{person}{Gloria Zen}, \bibinfo{person}{Bruno Lepri},
  \bibinfo{person}{Elisa Ricci}, {and} \bibinfo{person}{Oswald Lanz}.}
  \bibinfo{year}{2010}\natexlab{}.
\newblock \showarticletitle{Space Speaks: Towards Socially and Personality
  Aware Visual Surveillance}. In \bibinfo{booktitle}{\emph{Proceeding of the
  International Workshop on Multimodal Pervasive Video Analysis}}.
  \bibinfo{pages}{37--42}.
\newblock


\end{thebibliography}


\nobalance

\vspace{10mm}

\noindent{----------------}

\section*{Appendix}\label{sec:additional}

We present additional results and discussion regarding the behavior of our network and baselines under various conditions.

\vspace{-1mm}
\subsection*{\bf Varying History Window $H$}

In Tables \ref{tab:action_prediction_A1} and \ref{tab:action_prediction_A2}, we present results of training our networks with various history windows $H \in \{10,5,1\}$, using static and dynamic rollouts from the synthetic dataset initial layouts. As expected, lowering the history window size lowers performance of all networks, but \mgp{} either outperforms all baselines, or performs comparably to our strongest baseline Social Pooling. 

The strong performance of Social Pooling is due to the explicit spatial integration of neighbor information in a grid around the agent of interest, resulting in the implicit learning of sensory gating. Note that Social Pooling and Average Pooling networks have no data left out in their inputs - they each get the same data as our proposed network, since relative gaze and pose history is included in their visual history encoding. Furthermore, Social Pooling is made stronger by spatial binning instead of naive overall averaging. While our network must learn to appropriately weight each neighbor for pooling, Social Pooling may instead achieve the learning of attention by learning weighting over the grid. Moreover, propagating each grid's pooled messages forward results in a higher preservation of information than in our network. 

As expected, this advantage is visible in results at lower history window sizes in the static agent case, when it becomes harder to correctly learn neighbor weighting (Table \ref{tab:action_prediction_A1} in this document, Figure 6 in main submission). However, when neighbors are dynamic (Table \ref{tab:action_prediction_A2}), \mgp{} still outperforms all baselines, even at lower history windows. This is because the explicit sensory gating required to infer when to pay attention to a moving agent is advantageous over a static weighting learned over a grid, as in Social Pooling.

\begin{figure*}[h]
\centering
\includegraphics[width=0.9\linewidth]{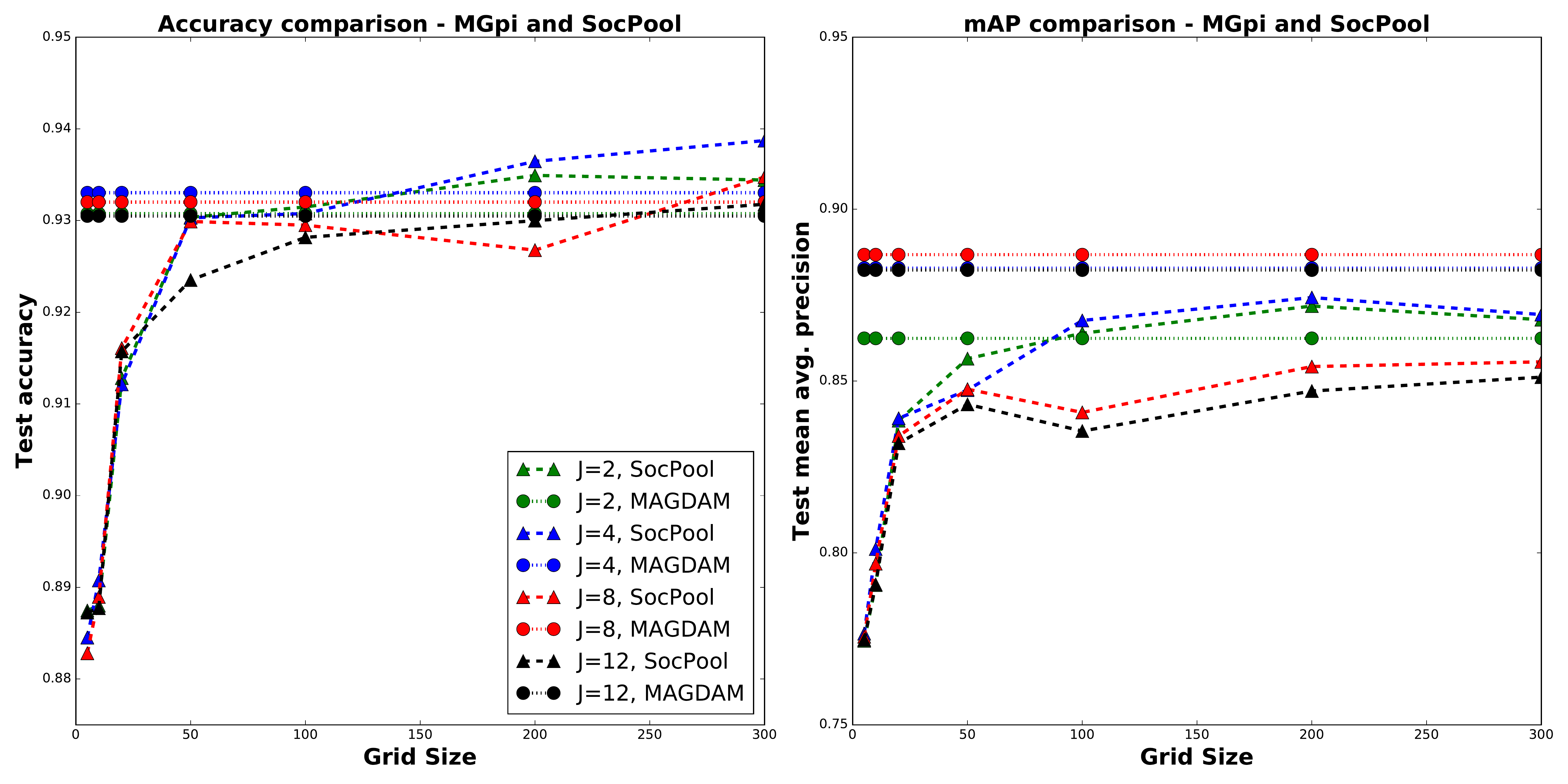}
\caption{{\bf Change in Social Pooling performance} with varying grid size, history window $H = 15$. \mgp{} performs similarly to \texttt{SOCPOOL} under the accuracy metric, and outperforms \texttt{SOCPOOL}  under the mean average precision metric, even with larger grid sizes.}
\label{fig:socpool_compare}
\end{figure*}

Further, we analyze the effect of varying number of grids in Social Pooling, at history $H = 15$. Results in Figure \ref{fig:socpool_compare} show that \mgp{} outperforms Social Pooling even with larger numbers of grids under the mean average precision metric, while performing similarly under the accuracy metric. Further, \mgp{} furnishes us with a natural way to infer group membership in real-world data (Section \ref{sec:expt_group}), which Social Pooling does not.

\vspace{-1mm}
\subsection*{{\bf Varying Initial Layouts}}

In Tables \ref{tab:action_prediction_B1}  and \ref{tab:action_prediction_B2} we present performance results using datasets generated from synthetic and real-world initial layouts in Synthetic \cite{Cristani2011}, Coffee Break \cite{Cristani2011} and Cocktail Party \cite{Zen2010} datasets. As expected, \mgp{} outperforms all baselines in the dynamic case, while performing comparably to Social Pooling in the static case. This comparison has been discussed in the previous subsection. 

\vspace{-1mm}
\subsection*{\bf Confusion Matrices}
\begin{figure*}[h]
\centering
\includegraphics[width=0.9\linewidth]{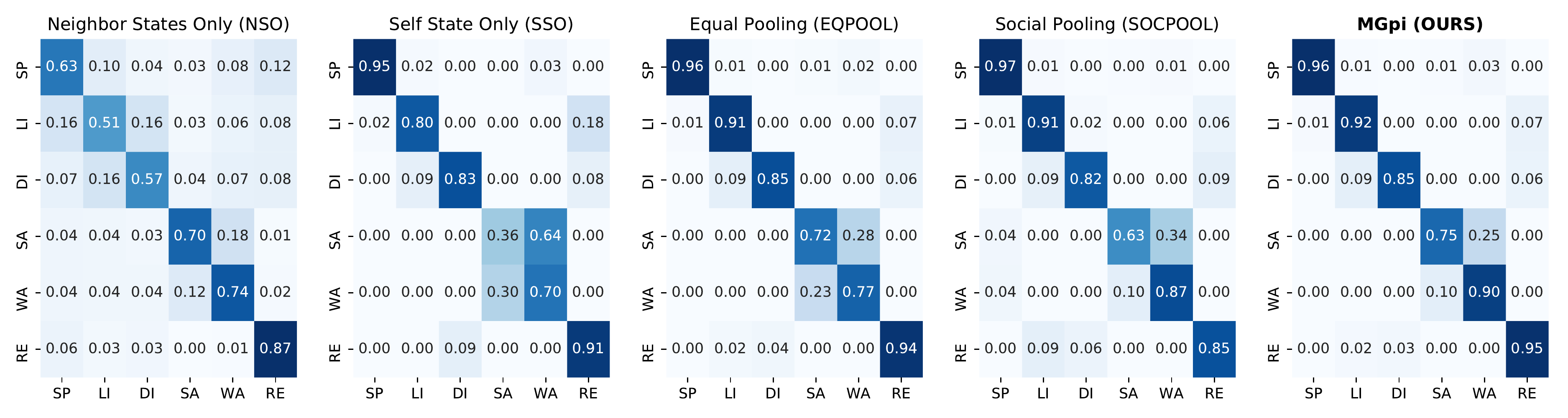}
\includegraphics[width=\linewidth]{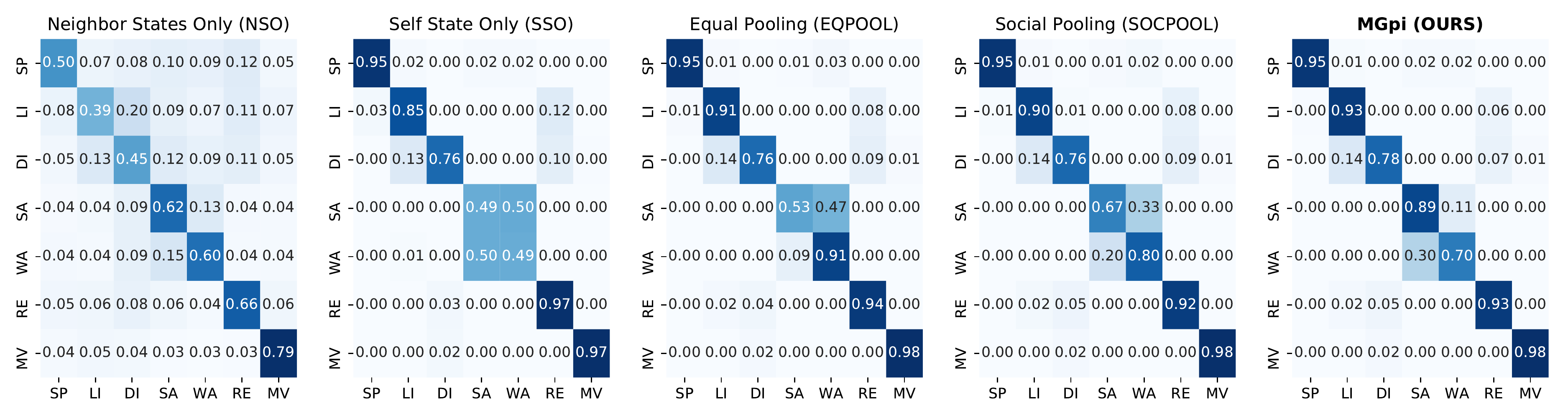}
\caption{{\bf Confusion matrices on action prediction:} Top ($6\times6$) - Static agents case. Bottom ($7\times7$) - Dynamic agents case. SP: Speaking, LI: Listening, DI: Distracted, SA: Strong Addressing, WA: Weak Addressing, RE: Responding, MV: Moving.}
\label{fig:cmat}
\end{figure*}

As mentioned in Section \ref{sec:sim}, confusion matrices of action prediction for all networks (Figure \ref{fig:cmat}) show that \texttt{NSO} and \texttt{SSO} work complementarily: \texttt{NSO} predicts better if agents make strong or weak addressing decisions by observing others. On the other hand, \texttt{SSO} performs better when observation of the self is necessary, \ie, in order to predict if agents keep speaking, listening, are distracted or responding. \mgp{} outperforms baselines for most actions, both in the static and dynamic case, with lower ambiguity.

\subsection*{\bf Learned Gating Outputs}

\begin{figure*}[h]
\centering
Synthetic:\;\;\;\;\;\;\;\;
\begin{subfigure}[]{0.4\textwidth}
    \centering
    \includegraphics[width=\linewidth]{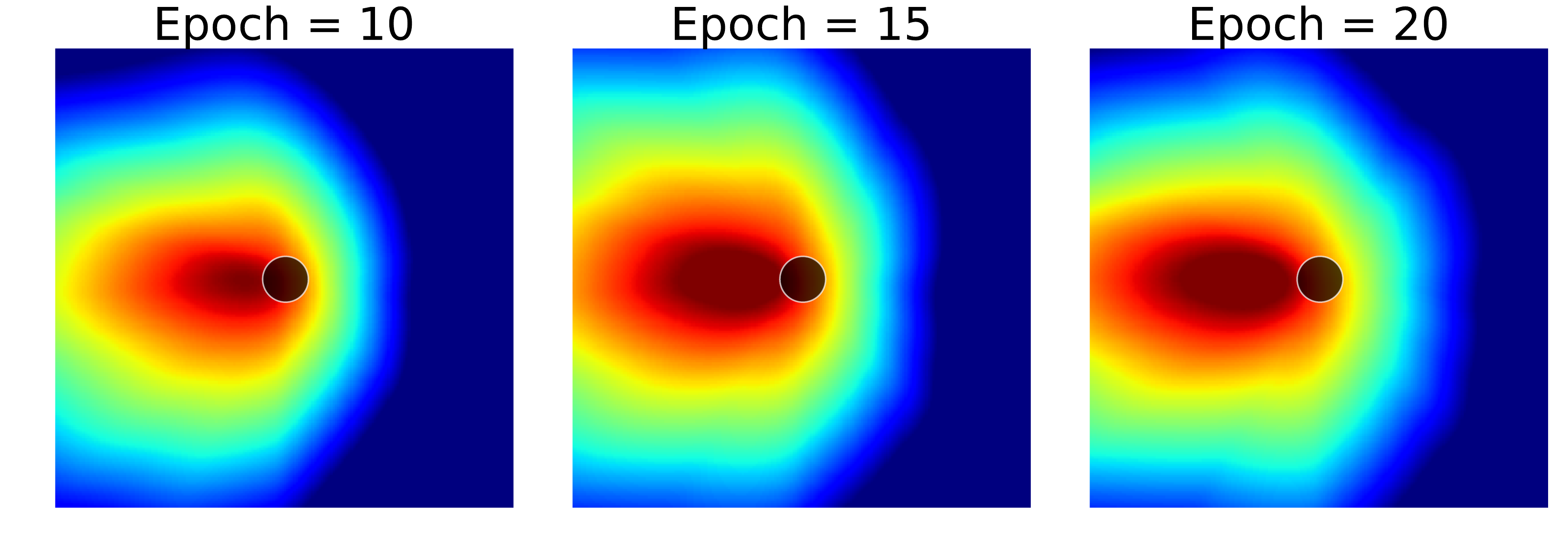}
\end{subfigure}
\hspace{2mm}
\begin{subfigure}[]{0.4\textwidth}
    \centering
    \includegraphics[width=\linewidth]{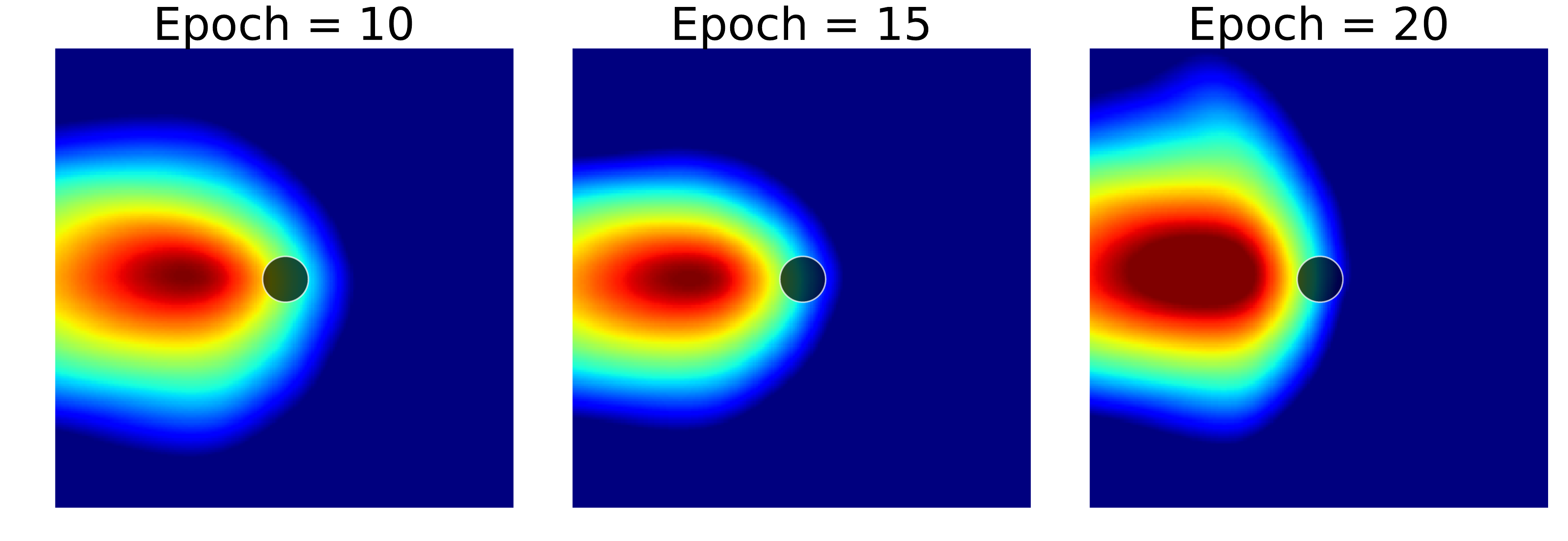}
\end{subfigure}

Coffee Break:\;\;
\begin{subfigure}[]{0.4\textwidth}
    \centering
    \includegraphics[width=\linewidth]{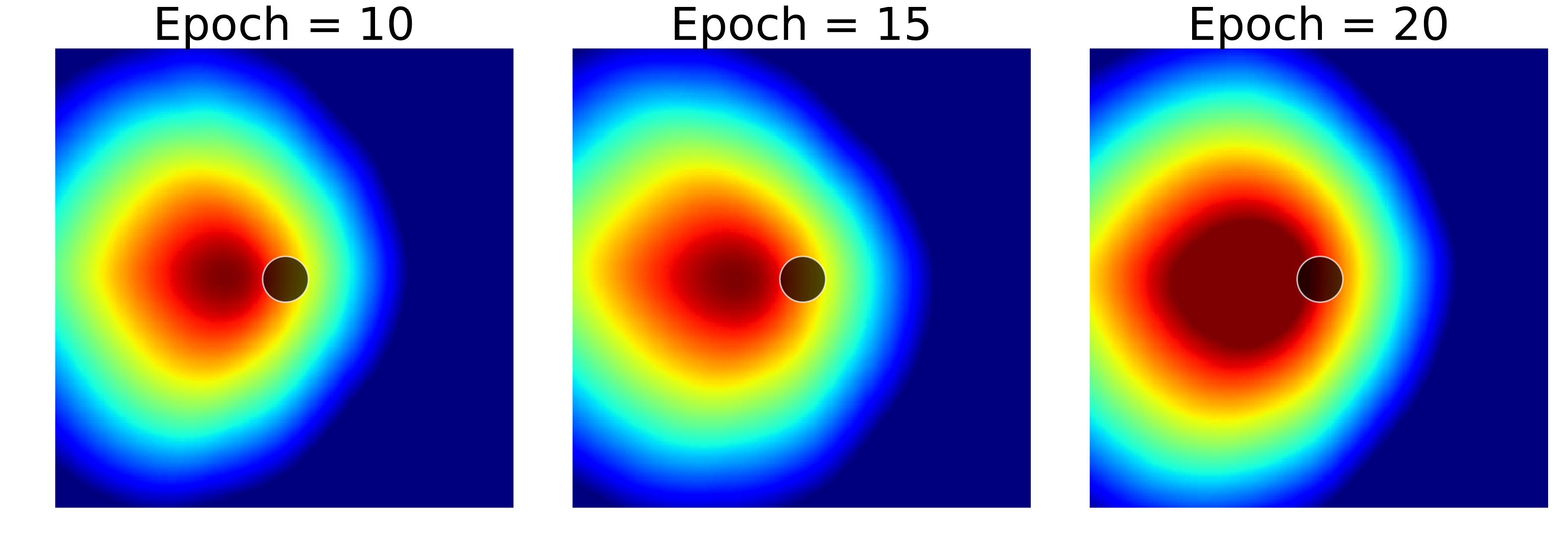}
\end{subfigure}
\hspace{2mm}
\begin{subfigure}[]{0.4\textwidth}
    \centering
    \includegraphics[width=\linewidth]{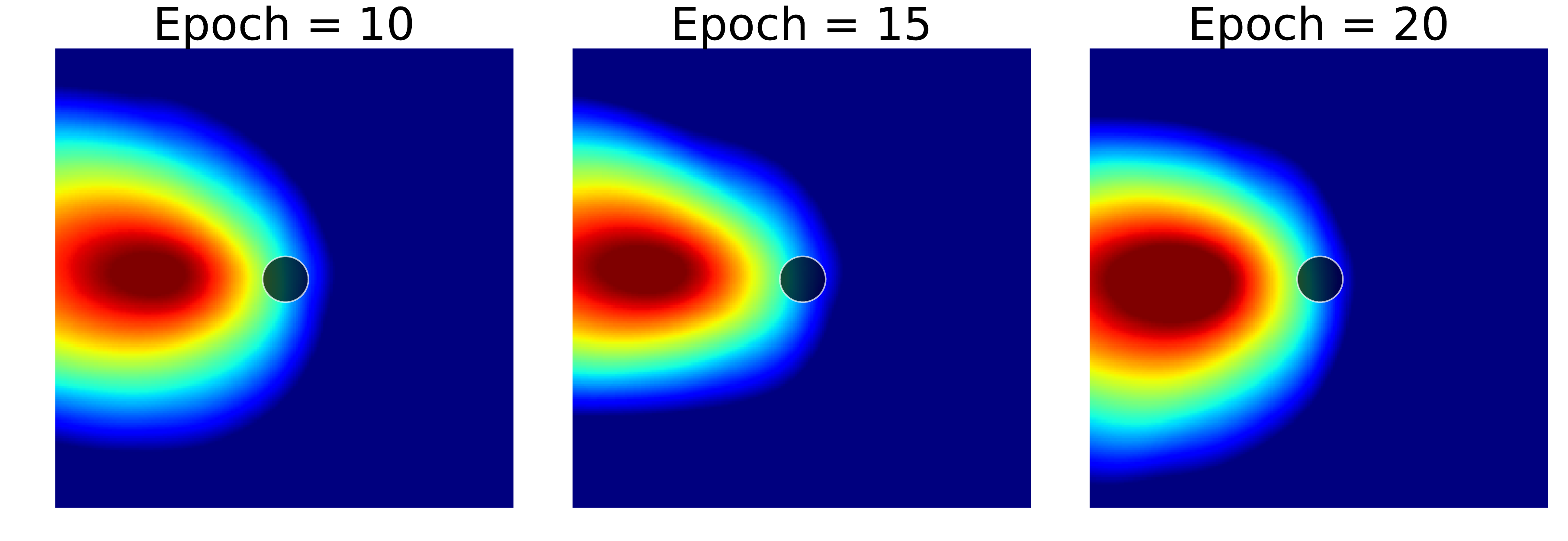}
\end{subfigure}

Cocktail Party:
\begin{subfigure}[]{0.4\textwidth}
    \centering
    \includegraphics[width=\linewidth]{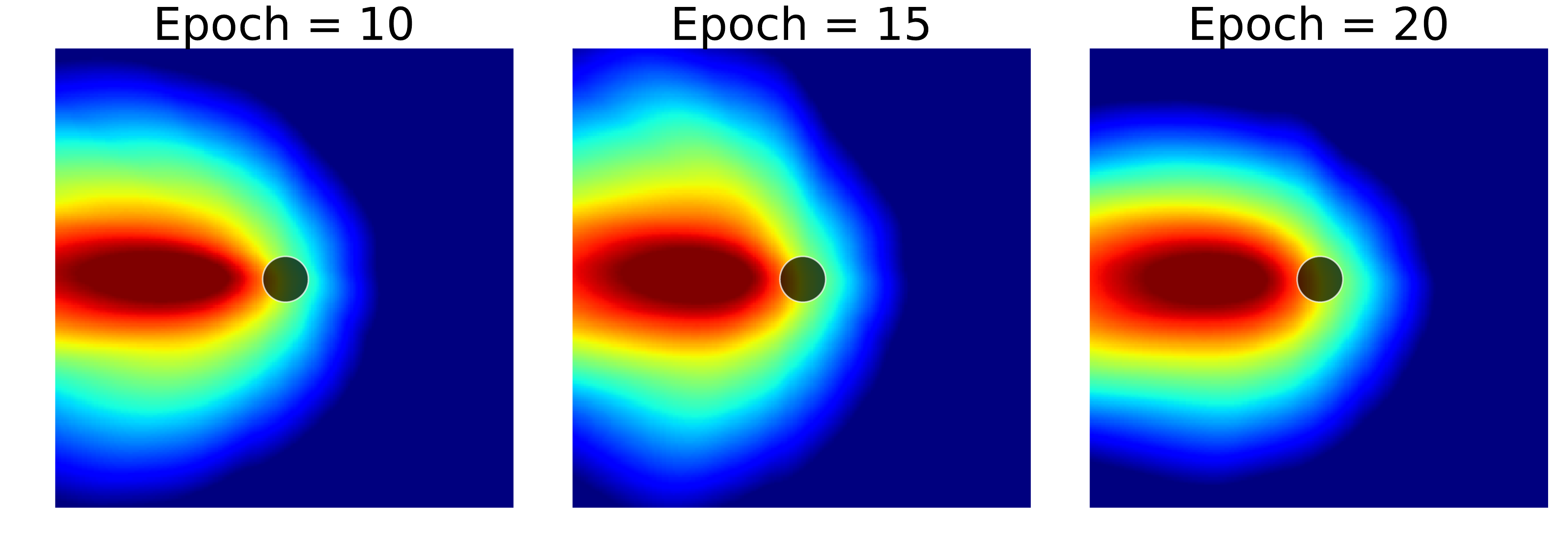}
    \caption{Static Agents Case}
\end{subfigure}
\hspace{2mm}
\begin{subfigure}[]{0.4\textwidth}
    \centering
    \includegraphics[width=\linewidth]{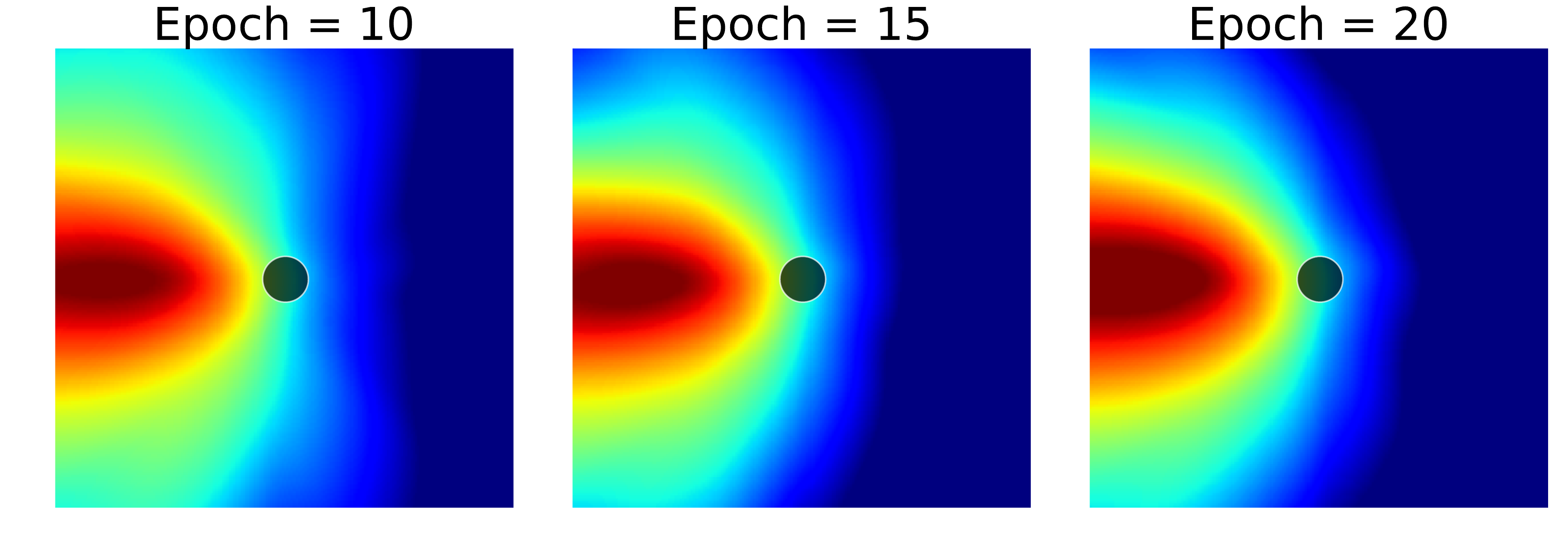}
    \caption{Dynamic Agents Case}
\end{subfigure}

\caption{{\bf Learned gating output for static agents (left) and dynamic agents (right) cases} for Synthetic, Coffee Break and Cocktail Party initial layouts, respectively. Each are displayed at training epochs 10, 15 and 20.}
\label{fig:kpm_compare}
\end{figure*}

We visualize outputs of the KPM-Gate for an agent $A^m$ at various surrounding locations of a neighbor agent $A^{n_i}$ while fixing relative gaze direction. We do so for all cases of initial layout source (Synthetic, Coffee Break or Cocktail Party) and conditions (static or dynamic agents) in Figure \ref{fig:kpm_compare}. Similar to Figure \ref{fig:visualization1}, the black circles represent $A^m$ looking in the $-x$-direction. $A^{n_i}$ is assumed to be looking in the $+x$-direction. Warmer colors indicate higher outputs from the gating function at those corresponding positions of $A^{n_i}$.

As described in Section \ref{sec:expt_group}, such outputs are used to infer group membership of a neighbor agent $A^{n_i}$ with respect to the agent under consideration $A^m$. As can be seen in Figure \ref{fig:kpm_compare}, our network learns appropriate gating in all cases by the $20^{\text{th}}$ epoch of training. Variations between the learned gating under various conditions leads to appropriate learning of the distance function $D(m,n_i)$ for clustering. This custom informative distance function leads to group detection performance comparable to state-of-the-art methods (Table 3).

\begin{table*}[!htbp]
\caption{Static Scenario: Action Prediction (mAP) for $H = 10, 5, 1$, Synthetic \cite{Cristani2011} initial layouts}
\centering
\scalebox{.7}{
\begin{tabular}{lcccc}
\toprule
      Model {\bf (mAP's, $H=10$)} &  $J=2$& 4&8&12 \\ 
     \midrule
     Neighbor States Only \hfill (\texttt{NSO}) & 0.65 & 0.68 & 0.66 & 0.65 \\
     Self States Only \hfill (\texttt{SSO}) & 0.72 & 0.72 & 0.72 & 0.72 \\
     Equal Pooling \hfill (\texttt{EQPOOL}) &0.85 & 0.85& 0.83 & 0.83\\
     Social Pooling \hfill (\texttt{SOCPOOL}) &0.85 & 0.85 & 0.84 & 0.83\\
     \midrule
     {\bf \mgp{} Network} & {\bf 0.87} & {\bf 0.88} & {\bf 0.86} & {\bf 0.85}\\
\bottomrule
\end{tabular}
}
\scalebox{.7}{
\begin{tabular}{lcccc}
\toprule
      Model {\bf (mAP's, $H=5$)} &  $J=2$& 4&8&12 \\ 
     \midrule
     Neighbor States Only \hfill (\texttt{NSO}) & 0.59 & 0.60 & 0.60 & 0.59 \\
     Self States Only \hfill (\texttt{SSO}) & 0.59 & 0.59 & 0.59 & 0.59 \\
     Equal Pooling \hfill (\texttt{EQPOOL}) &0.80 & 0.79& 0.76 & 0.76\\
     Social Pooling \hfill (\texttt{SOCPOOL}) &{\bf 0.80} & {\bf 0.80} & {\bf 0.80} & 0.79\\
     \midrule
     {\bf \mgp{} Network} & 0.79 & {\bf 0.80} & {\bf 0.80} & {\bf 0.80}\\
\bottomrule
\end{tabular}
}
\scalebox{.7}{
\begin{tabular}{lcccc}
\toprule
      Model {\bf (mAP's, $H=1$)} &  $J=2$& 4&8&12 \\ 
     \midrule
     Neighbor States Only \hfill (\texttt{NSO}) & 0.48 & 0.48 & 0.49 & 0.47 \\
     Self States Only \hfill (\texttt{SSO}) & 0.49 & 0.49 & 0.49 & 0.49 \\
     Equal Pooling \hfill (\texttt{EQPOOL}) &0.64 & 0.63& 0.62 & 0.60\\
     Social Pooling \hfill (\texttt{SOCPOOL}) &{\bf 0.69} & {\bf 0.70} & {\bf 0.69} & {\bf 0.69}\\
     \midrule
     {\bf \mgp{} Network} &  {0.68} & {0.68} & {0.67} & {0.66}\\
\bottomrule
\end{tabular}
}\label{tab:action_prediction_A1}
\end{table*}

\begin{table*}[!htbp]
\caption{Dynamic Scenario: Action Prediction (mAP) for $H = 10, 5, 1$, Synthetic \cite{Cristani2011} initial layouts}
\centering
\scalebox{.7}{
\begin{tabular}{lcccc}
\toprule
      Model {\bf (mAP's, $H=10$)} &  $J=2$& 4&8&12 \\ 
     \midrule
     Neighbor States Only \hfill (\texttt{NSO}) & 0.55 & 0.58 & 0.62 & 0.62 \\
     Self States Only \hfill (\texttt{SSO}) & 0.73 & 0.73 & 0.73 & 0.73 \\
     Equal Pooling \hfill (\texttt{EQPOOL}) &0.82 & 0.83& 0.82 & 0.81\\
     Social Pooling \hfill (\texttt{SOCPOOL}) &0.82 & 0.83 & 0.83 & 0.83\\
     \midrule
     {\bf \mgp{} Network} & {\bf 0.83} & {\bf 0.85} & {\bf 0.85} & {\bf 0.86}\\
\bottomrule
\end{tabular}
}
\scalebox{.7}{
\begin{tabular}{lcccc}
\toprule
      Model {\bf (mAP's, $H=5$)} &  $J=2$& 4&8&12 \\ 
     \midrule
     Neighbor States Only \hfill (\texttt{NSO}) & 0.50 & 0.55 & 0.58 & 0.58 \\
     Self States Only \hfill (\texttt{SSO}) & 0.63 & 0.63 & 0.63 & 0.63 \\
     Equal Pooling \hfill (\texttt{EQPOOL}) &0.75 & 0.75& 0.74 & 0.75\\
     Social Pooling \hfill (\texttt{SOCPOOL}) &{\bf 0.76} & 0.77 & 0.78 & 0.78\\
     \midrule
     {\bf \mgp{} Network} & {\bf 0.76} & {\bf 0.78} & {\bf 0.79} & {\bf 0.79}\\
\bottomrule
\end{tabular}
}
\scalebox{.7}{
\begin{tabular}{lcccc}
\toprule
      Model {\bf (mAP's, $H=1$)} &  $J=2$& 4&8&12 \\ 
     \midrule
     Neighbor States Only \hfill (\texttt{NSO}) & 0.37 & 0.41 & 0.44 & 0.43 \\
     Self States Only \hfill (\texttt{SSO}) & 0.55 & 0.55 & 0.55 & 0.55 \\
     Equal Pooling \hfill (\texttt{EQPOOL}) &0.62 & 0.62& 0.61 & 0.61\\
     Social Pooling \hfill (\texttt{SOCPOOL}) &{0.63} & {0.64} & {0.64} & {0.64}\\
     \midrule
     {\bf \mgp{} Network} &  {\bf 0.64} & {\bf 0.65} & {\bf 0.66} & {\bf 0.66}\\
\bottomrule
\end{tabular}
}
\label{tab:action_prediction_A2}
\end{table*}

\begin{table*}[!htbp]
\caption{Static Scenario: Action Prediction (mAP) for $H = 15$, Synthetic \cite{Cristani2011}, Coffee Break \cite{Cristani2011}, Cocktail Party \cite{Zen2010} initial layouts}
\centering
\scalebox{.7}{
\begin{tabular}{lcccc}
\toprule
      Model {\bf (mAP's, Synthetic)} &  $J=2$& 4&8&12 \\ 
     \midrule
     Neighbor States Only \hfill (\texttt{NSO}) & 0.67 & 0.69 & 0.70 & 0.68 \\
     Self State Only \hfill (\texttt{SSO}) & 0.76 & 0.76 & 0.76 & 0.76 \\
     Equal Pooling \hfill (\texttt{EQPOOL}) &0.87 & 0.86& 0.85 & 0.85\\
     Social Pooling \hfill (\texttt{SOCPOOL}) &0.85 & 0.84 & 0.84 & 0.84\\
     \midrule
     {\bf \mgp{} Network} & {\bf 0.88} & {\bf 0.88} & {\bf 0.89} & {\bf 0.88}\\
\bottomrule
\end{tabular}
}
\scalebox{.7}{
\begin{tabular}{lcccc}
\toprule
      Model {\bf (mAP's, Coffee Break)} &  $J=2$& 4&8&12 \\ 
     \midrule
     Neighbor States Only \hfill (\texttt{NSO}) & 0.71 & 0.72 & 0.71 & 0.70 \\
     Self States Only \hfill (\texttt{SSO}) & 0.75 & 0.75 & 0.75 & 0.75 \\
     Equal Pooling \hfill (\texttt{EQPOOL}) &0.89 & 0.88& 0.87 & 0.86\\
     Social Pooling \hfill (\texttt{SOCPOOL}) &{\bf 0.91} & {\bf 0.92} & {\bf 0.91} & {\bf 0.91}\\
     \midrule
     {\bf \mgp{} Network} & {0.90} & {0.91} & {\bf 0.91} & {\bf 0.91}\\
\bottomrule
\end{tabular}
}
\scalebox{.7}{
\begin{tabular}{lcccc}
\toprule
      Model {\bf (mAP's, Cocktail Party)} &  $J=2$& 4&8&12 \\ 
     \midrule
     Neighbor States Only \hfill (\texttt{NSO}) & 0.72 & 0.79 & 0.79 & 0.79 \\
     Self States Only \hfill (\texttt{SSO}) & 0.73 & 0.73 & 0.73 & 0.73 \\
     Equal Pooling \hfill (\texttt{EQPOOL}) &0.91 & 0.94& 0.94 & 0.94\\
     Social Pooling \hfill (\texttt{SOCPOOL}) &{\bf 0.92} & {\bf 0.95} & {0.94} & 0.94\\
     \midrule
     {\bf \mgp{} Network} & {\bf 0.92} & {\bf 0.95} & {\bf 0.95} & {\bf 0.95}\\
\bottomrule
\end{tabular}
}
\label{tab:action_prediction_B1} 
\end{table*}

\begin{table*}[h]
\caption{Dynamic Scenario: Action Prediction (mAP) for $H = 15$, Synthetic \cite{Cristani2011}, Coffee Break \cite{Cristani2011}, Cocktail Party \cite{Zen2010} initial layouts}
\centering
\scalebox{.7}{
\begin{tabular}{lcccc}
\toprule
      Model {\bf (mAP's)} &  $J=2$& 4&8&12 \\ 
     \midrule
     Neighbor States Only \hfill (\texttt{NSO}) & 0.57 & 0.62 & 0.65 & 0.64 \\
     Self State Only \hfill (\texttt{SSO}) & 0.78 & 0.78 & 0.78 & 0.78 \\
     Equal Pooling \hfill (\texttt{EQPOOL}) &0.85 & 0.86 & 0.85 & 0.85\\
     Social Pooling \hfill (\texttt{SOCPOOL}) &0.85 & 0.85 & 0.86 & 0.85\\
     \midrule
     {\bf \mgp{} Network} & {\bf 0.87} & {\bf 0.87} & {\bf 0.88} & {\bf 0.88}\\
\bottomrule
\end{tabular}
}
\scalebox{.7}{
\begin{tabular}{lcccc}
\toprule
      Model {\bf (mAP's, Coffee Break)} &  $J=2$& 4&8&12 \\ 
     \midrule
     Neighbor States Only \hfill (\texttt{NSO}) & 0.60 & 0.65 & 0.68 & 0.68 \\
     Self States Only \hfill (\texttt{SSO}) & 0.78 & 0.78 & 0.78 & 0.78 \\
     Equal Pooling \hfill (\texttt{EQPOOL}) &0.86 & 0.87& 0.87 & 0.86\\
     Social Pooling \hfill (\texttt{SOCPOOL}) &0.86 & 0.87 & 0.87 & 0.87\\
     \midrule
     {\bf \mgp{} Network} & {\bf 0.87} & {\bf 0.88} & {\bf 0.89} & {\bf 0.89}\\
\bottomrule
\end{tabular}
}
\scalebox{.7}{
\begin{tabular}{lcccc}
\toprule
      Model {\bf (mAP's, Cocktail Party)} &  $J=2$& 4&8&12 \\ 
     \midrule
     Neighbor States Only \hfill (\texttt{NSO}) & 0.59 & 0.73 & 0.75 & 0.74 \\
     Self States Only \hfill (\texttt{SSO}) & 0.78 & 0.78 & 0.78 & 0.78 \\
     Equal Pooling \hfill (\texttt{EQPOOL}) &0.89 & 0.91& 0.91 & 0.91\\
     Social Pooling \hfill (\texttt{SOCPOOL}) &{0.87} & {0.90} & {0.90} & 0.90\\
     \midrule
     {\bf \mgp{} Network} & {\bf 0.89} & {\bf 0.92} & {\bf 0.93} & {\bf 0.92}\\
\bottomrule
\end{tabular}
}
\label{tab:action_prediction_B2} 
\end{table*}

\clearpage

\end{document}